\documentclass{article}

\usepackage{arxiv}

\usepackage[utf8]{inputenc} 
\usepackage[T1]{fontenc}    
\usepackage{hyperref}       
\usepackage{url}            
\usepackage{booktabs}       
\usepackage{amsfonts}       
\usepackage{nicefrac}       
\usepackage{microtype}      
\usepackage{lipsum}		
\usepackage{graphicx}
\usepackage{natbib}
\usepackage{doi}

\title{Level of agreement between emotions generated by artificial intelligence and human evaluation: A methodological proposal}

\date{October 10, 2024}	

\author{ \href{https://orcid.org/0000-0002-5389-7590}{\includegraphics[scale=0.06]{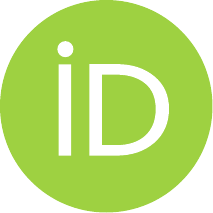}\hspace{1mm}Miguel Carrasco}\thanks{Corresponding author: personal webpage: \href{http://mlacarrasco.github.io}{http://mlacarrasco.github.io}} \\
	Facultad de Ingeniería y Ciencias\\
	Universidad Adolfo Ibañez\\
	Santiago, Chile \\
	\texttt{miguel.carrasco@uai.cl} \\
	\And
	\href{https://orcid.org/0000-0002-9017-3196}{\includegraphics[scale=0.06]{orcid.pdf}\hspace{1mm}César González-Martín} \thanks{Corresponding author}\\
	Faculty of Education Sciences and Psychology\\
	University of Cordoba\\
	\texttt{cesar.gonzalez@uco.es} \\
    \And
	\href{https://orcid.org/0000-0002-9601-4305}{\includegraphics[scale=0.06]{orcid.pdf}\hspace{1mm}Sonia Navajas-Torrente} \\
	Faculty of Law, Economics and Business\\
	University of Cordoba\\
   \And
	Raúl Dastres \\
	Facultad de Ingeniería y Ciencias\\
	Universidad Adolfo Ibañez\\
	Santiago, Chile \\
}

\renewcommand{\shorttitle}{Article}

\hypersetup{
pdftitle={Level of agreement between emotions generated by artificial intelligence and human evaluation: A methodological proposal},
pdfsubject={q-bio.NC, q-bio.QM},
pdfauthor={Miguel Carrasco, César González-Martín, Sonia Navajas-Torrente, Raúl Dastres},
pdfkeywords={Agreement, Emotion, Generative Neural Networks},
}

\begin{document}
\maketitle

\begin{abstract}
	Images are capable of conveying emotions, but emotional experience is highly subjective. Advances in artificial intelligence have enabled the generation of images based on emotional descriptions. However, the level of agreement between the generative images and human emotional responses has not yet been evaluated. To address this, 20 artistic landscapes were generated using StyleGAN2-ADA. Four variants evoking positive emotions (contentment, amusement) and negative emotions (fear, sadness) were created for each image, resulting in 80 pictures. An online questionnaire was designed using this material, in which 61 observers classified the generated images. Statistical analyses were performed on the collected data to determine the level of agreement among participants, between the observers' responses, and the AI's generated emotions. A generally good level of agreement was found, with better results for negative emotions. However, the study confirms the subjectivity inherent in emotional evaluation.
\end{abstract}


\section{Introduction}
An image serves as a means of communication, conveying a message capable of evoking emotions based on the intention behind its creation \cite{lyu_cognition_2021}. To ensure the accurate interpretation of the message by the observer, it is essential to implement a well-designed visual strategy. This strategy serves as a channel to elicit both conscious and unconscious emotional reactions, which manifest physiologically \cite{li_physiological-signal-based_2021} and psychologically \cite{hess_signal_2016,lin_review_2023,sharma_computerized_2021}. However, one of the main challenges in studying emotions is the subjective nature of emotional responses to experiences, which can vary significantly between individuals\cite{zhao_affective_2022}. Therefore, reaching a significant agreement between individuals is complex \cite{eser_comparison_2022}, and even more so between generative artificial intelligence and humans.

In addition to subjectivity, other factors affect the correct experience that produces the emotional response, such as the observer's socio-cultural context \cite{ali_high-level_2017}, their experience  \cite{joshi_aesthetics_2011,lim_cultural_2016,redies_global_2020,russell_cross-cultural_2017}, the temporal evolution of the emotion or the location of the image \cite{ionescu_computational_2022}, which can produce unexpected reactions \cite{peng_mixed_2015} contrary to the initial purpose in visual creation.

These factors pose a challenge for emotion categorization. For this reason, psychology has developed Categorical Emotion States (CES) or discrete models, which identify basic emotions, as proposed by Ekman or Mikels. In contrast, the multi-dimensional model (DES) \citet{wood_comparison_2018} categorizes emotions based on valence, which defines pleasure; arousal, ranging from excitement to calm; and dominance, the degree of control. In this model, emotions are often binarized as positive or negative, although sometimes a neutral category is included \cite{wang_systematic_2022,zhao_affective_2022}.

The complexity of the relationship between visual objects and emotions, along with the ongoing quest to understand emotional processes, plays a crucial role in human cognition, communication, and behaviour \cite{lin_review_2023}. Due to the neurophysiological responses triggered by everyday situations or mental processes such as memories, imagination, or beliefs \cite{li_physiological-signal-based_2021}, this has become a broad field of neuroscience research, focused on the analysis and recognition of emotions \cite{suhaimi_eeg-based_2020}, as well as in psychology \cite{egger_emotion_2019,lin_review_2023}, education \cite{imani_survey_2019}, and health \cite{hasnul_electrocardiogram-based_2021}. A wide range of methodologies has been employed \cite{khare_emotion_2024}, including speech analysis \cite{al-dujaili_speech_2023}, facial expressions \cite{leong_facial_2023}, body movement \cite{ebdali_takalloo_overview_2022}, thermal measurements \cite{fardian_thermography_2022}, text analysis \cite{kusal_review_2022}, as well as movies \cite{almeida_emotion_2021}, music \cite{han_survey_2022}, and multimodal approaches \cite{pan_review_2023}.

In addition to these areas, the field of computer science, especially computer vision, has taken an interest in art as an object of study for the analysis of emotions, in visual emotion analysis \cite{zhao_emotional_2020}, emotion recognition \cite{ahmed_systematic_2023,dzedzickis_human_2020} or Affective image content analysis \cite{ali_high-level_2017,zhao_affective_2022}, where the denotative elements of the image, known as low-level, local or handcrafted features \cite{bianco_multitask_2019,cetinic_fine-tuning_2018,dewan_image_2020,wang_review_2019,zhao_exploring_2014}, such as textures \cite{abry_when_2013,guo_application_2022,kelishadrokhi_innovative_2023,liu_power_2018}, shapes \cite{lu_shape_2012} or colour \cite{Priya_affective_2020,bianco_multitask_2019,kang_method_2018,peng_framework_2014}, are identified and analyzed. Semantics are typically referred to as high-level or global features \cite{ali_high-level_2017,li_object_2014,machajdik_affective_2010,redies_global_2020,tu_unsupervised_2023,zhao_pdanet_2019,zhao_affective_2022}. In this context, research has focused on the analysis and classification of aesthetic aspects \cite{fekete_vienna_2022,joshi_aesthetics_2011}, places \cite{ali_high-level_2017}, and emphasis and harmony \cite{zhao_exploring_2014}, which involves an analysis of several characteristics of the image. In some cases, the relationships between attributes or compositions have been studied, known as mid-level or semi-local features \cite{fernando_mining_2014,gordo_supervised_2015,machajdik_affective_2010,zhu_dependency_2017}.

Thus, studies can be found by analyzing abstract art \cite{alameda-pineda_recognizing_2016,he_emotion_2018,sartori_whos_2015} and oriental art \cite{hung_study_2018,li_emotion_2021,tian_kaokore_2020,wang_aed_2021,zhang_research_2021}, cubist art \cite{ginosar_detecting_2014}, figurative art \cite{hagtvedt_perception_2008}, artwork from various cultures \cite{stamatopoulou_feeling_2017}, photography \cite{joshi_aesthetics_2011,stampfl_role_2023,yang_distinguishing_2019}, and public art \cite{tian_novel_2022}, painting in general \cite{bianco_multitask_2019,kang_method_2018,tashu_attention-based_2021}, drawing \cite{yin_object_2016}, comics \cite{she_learning_2019}, portraits of different artistic styles \cite{yang_distinguishing_2019} and techniques \cite{guo_application_2022}, or investigating whether there are differences between disciplines \cite{yang_distinguishing_2019}, or using the title of the work or the author \cite{sartori_whos_2015,tashu_attention-based_2021}.

Nowadays, in addition to the field of automatic emotion recognition, there is a growing development of generative artificial intelligence (AI) capable of creating images based on emotional input (e.g., prompts) \cite{sivasathiya_emotion-aware_2024}. This development has highlighted the need for additional processing to validate the generated content \cite{hajarolasvadi_generative_2020}. Validation can be conducted by considering various aspects of the images, including visual elements such as formats, color, textures, and connotative elements like meaning or intrinsic emotions. Our research focuses on the emotional validation of images generated by AI. Thus, the research hypothesis investigates whether images created through generative processes with a specified emotion align with human emotional responses to a significant degree.

Given this need, and to the best of our knowledge, no previous studies are conducting statistical analyses on the level of agreement between emotions generated by generative artificial intelligence and human judgment, this research proposes a methodology to address this issue. It has been developed in three phases: In the data preparation phase, the Artemis dataset was used to train the generative model StyleGAN2-ADA. In the modelling phase, 20 landscape images were generated, with four variants for each image—two expressing positive emotions and two expressing negative emotions—resulting in a total of 80 images. Finally, in the evaluation phase, an online questionnaire was designed using this set of images, where 61 individuals classified the images according to their emotions. Subsequently, various statistical analyses were conducted to establish the degree of agreement among individuals, including Krippendorff's Alpha, the mode of the individuals and the AI using precision, recall, F1-Score, and Fisher's test, and proportion analysis with Jaccard's index and Fisher's test.

In summary, this research proposes the following results and their validation. The following description does not constitute a methodology but rather outlines the results, as the methodology will be reviewed in subsequent sections:

\begin{itemize} 
\item Construction of a dataset composed of 80 generated images, artificially categorized into four emotional groups, accomplished using the StyleGAN-ADA2 tool. 
\item Evaluation among participants for each image to establish a baseline for comparison. 
\item Comparison between the mode of participants' responses and the emotions generated by AI. 
\item Evaluation of each individual's response to emotions generated by AI. 
\item Analysis of the proportions that align with AI-generated emotions. 
\item Evaluation of the hypothesis, contributing to the field of generative AI, as to our knowledge, no prior studies are measuring the consistency and level of agreement between AI-generated content and human perception.
\end{itemize}

\section{Background}
Image generation using computational techniques has experienced significant advancements in recent years. Traditional methods, such as rule-based image processing and image synthesis techniques, have evolved into more sophisticated approaches that rely on machine learning and convolutional neural networks. This transformation is largely attributed to the rapid progress in AI, driven by the continuous generation of large-scale data. Consequently, this advancement has led to the development of considerably more accurate and reliable AI models capable of generating images, that are practically indistinguishable from authentic photographs or paintings.

To examine and understand the computational techniques used in image generation, this section focuses on the current state of these techniques, with particular emphasis on generating artistic images. The analysis will be conducted through a comprehensive review of scientific and technical literature, ranging from traditional methods to the most innovative approaches based on machine learning and neural networks.

\subsection{Recurrent Neural Networks (RNN) and Convolutional Neural Networks (CNN)}
Recurrent Neural Networks (RNN) are one of the types of networks that have stood out in image generation. This type of network has proven useful for completing images from a section of them. The model presented by Google DeepMind in 2016, called PixelRNN \cite{van_den_oord_pixel_2016}, manages to understand the generality of pixel interdependence, being able to predict missing pixels in an image by receiving only a part of it. This type of network has also been used for generating images from natural language text descriptions~\cite{mansimov_generating_2015}. This study proposes an attention-based approach, where the model iteratively draws while focusing on the key words of the given description. The results obtained achieve the generation of higher resolution images than those obtained with other approaches, and generate images with a novel scene composition.

A type of neural network perhaps more widely used than Recurrent Networks are Convolutional Neural Networks or CNNs. Using this architecture, it has been possible to generate three-dimensional images of objects from different perspectives, as in the case of the model presented by \citet{dosovitskiy_learning_2015}, where by training convolutional networks, they managed to generate images of chairs from different perspectives. Another example of the use of Convolutional Networks is seen in a study conducted by Google DeepMind \cite{van_den_oord_conditional_2016}, where conditional image generation is explored using convolutional networks through PixelCNN. This model is capable of generating a variety of portraits of the same person using different facial expressions, poses, and lighting conditions. Its results are on par with PixelRNN, but it achieves this at a much lower computational cost.

The use of RNNs and CNNs is not mutually exclusive. For instance, a study on a Recurrent Convolutional Encoder-Decoder architecture \cite{yang_weakly-supervised_2015} demonstrates this integration. In this approach, convolutional networks handle both encoding and decoding, while a recurrent network manages object rotation. This combined strategy effectively synthesizes unseen versions of three-dimensional objects, enabling the generation of images of faces or chairs from various angles.

\subsection{Variational Auto-encoders (VAEs)}
The architecture known as Variational Auto-encoders has great utility as a generative model. Numerous studies have demonstrated the use of this architecture for image generation, an example of this is the so-called Deep Recurrent Attentive Writer or DRAW \cite{gregor_draw_2015}. This model uses a neural network that combines a spatial attention mechanism, mimicking the way human eyes move to focus on objects, with a self-encoding framework that allows the construction of complex images. DRAW has managed to obtain realistic results in the generation of various types of images, such as photographs of house numbers, in addition to the classic handwritten numbers.

Another example of the use of variational autoencoders is PixelVAE \cite{gulrajani_pixelvae_2016}. This model has an autoregressive decoder based on PixelCNN \cite{van_den_oord_conditional_2016}, but unlike this, it requires a smaller number of computationally expensive autoregression layers, making it more efficient. Additionally, this model manages to learn latent codes that are more compressed than a traditional VAE, while still capturing most of the non-trivial structures. This model presents comparable and competitive results, depending on the dataset, with other state-of-the-art methods. 

PixelVAE++ \cite{sadeghi_pixelvae_2019}, a generative model based on PixelVAE, is a VAE with three types of latent variables and, unlike PixelVAE, uses a PixelCNN++ network as a decoder. This model also presents a renewed architecture, where part of the decoder is reused as an encoder. PixelVAE++ presents superior results on the CIFAR-10 dataset compared to other latent variable models.

\citet{maalo_e_biva_2019} present BIVA, a bidirectional interface VAE, characterized by a ``skip-connected'' generative model and an inference network formed by a bidirectional stochastic inference path. This approach achieves good results, on par with other approaches, but proves to be useful not only for image generation but also for anomaly detection.

\subsection{Generative Adversarial Networks (GANs)}
Generative models have a long history. However, it will not be until the development of Deep Learning, that models will achieve significant advances \cite{cao_comprehensive_2023}. 
Introduced by \citet{goodfellow_generative_2014}, generative adversarial networks (GANs) have achieved important results in image processing and have attracted interest from the academic and industrial worlds in various fields of research and applications \cite{alqahtani_applications_2021,wang_state---art_2020}. The most relevant variants for image generation in GANs are conditional (cGANs), deep convolutional (DCGANs), and recurrent adversarial networks \cite{shahriar_gan_2022}.

As proposed by \citet{mirza_conditional_2014}, cGANs allow the generation of images conditioned on an additional input, which could be a class label or a reference image. Over the years, new algorithms based on projections \cite{miyato_cgans_2018} have emerged, considerably improving the performance of trained generators. \citet{odena_conditional_2017} proposed a variant of GANs, called Auxiliary Classifier GANs (AC-GANs). In this new variant, each generated sample has its corresponding class label in addition to the input noise, which is used together to generate an image. The discriminator gives both inputs a probability distribution, which means that the objective function has two parts: the log probability that the source is correct, and the log probability that the class is correct. AC-GANs achieve excellent results compared with traditional cGANs.

The ability to condition GANs on a second input opened the door to countless possibilities for this architecture, from something as basic as training the same model to generate cats and dogs, to something that seems futuristic as generating an image from a natural-language text description. An example is the generation of more realistic images from sketches \cite{kuriakose_synthesizing_2020,liu_sketch--art_2020,liu_auto-painter_2018,philip_face_2017}.

Meanwhile, \citet{radford_unsupervised_2015} presented Deep Convolutional GANs (DCGANs), a class of GANs that introduce upscaling convolutional layers between the input and output images of the generator, as well as using convolutional networks in the discriminator to determine whether the image is real or fake. One of the indications for stable DCGANs is that pooling layers should be replaced by scaled convolutions or ``strided convolutions'' in the discriminator and by scaled fractional convolutions or ``fractional-strided convolutions''. This alteration of GANs considerably stabilizes the training and generates higher-quality and higher-resolution images than traditional GANs. Given the success of convolutional neural networks (CNNs) in image and video classification in recent years, DCGAN remains a suitable architecture for image-generation applications \cite{shahriar_gan_2022}. The works of \cite{elgammal_can_2017,tian_kaokore_2020} using Style-based can be highlighted.

Style-based architectures in GANs are based on deconstructing high-level feature attributes from low-level features. An example of this type of architecture is StyleGAN \cite{karras_style-based_2021}, a variant of GANs presented by NVIDIA, which is inspired by the style transfer literature. Its architecture differs from that of traditional GANs by skipping the latent code input layer instead of starting with a learned constant. Given a latent code, a nonlinear network produces a version of a generative image found in a latent space, which then controls the generator through adaptive instance normalization (AdaIN) in each convolutional layer. This revolutionized image generation is owing to its diversity and high realistic capacity \cite{bandi_power_2023}.

This architecture has received updates, such as StyleGAN2 \cite{karras_analyzing_2020}, implementing progressive growth and regularizing the generator to drive good conditioning in the mapping of latent codes to images. As an alternative, StyleGAN2-ADA \cite{karras_training_2020} was released, where an adaptive discriminator augmentation mechanism was implemented that stabilizes training when training with a limited amount of data. These additions yield good results even with little data. Finally, the latest update, called StyleGAN3 
\cite{zhang_styleswin_2022}, implements small changes to the architecture to ensure that unimportant information does not leak into the hierarchical synthesis process. The resulting networks are on par with the StyleGAN2 FID results but vary completely in their internal interpretation and are completely invariant to translation, even at the sub-pixel scale. The results of this latest version of StyleGAN were better for models focused on videos and animations.

Another style-based GAN architecture was presented by Microsoft StyleSwin \cite{zhang_styleswin_2022}. This variant explores the option of building a generative adversarial network model using pure transformers, in which the proposed generator adopts the Swin transform. This model is scalable to high-resolution images and achieves excellent results with the FFHQ-1024 and CelebA-HQ 1024 datasets.

On the other hand, one of the challenges that have been addressed in recent years is the generation of new artistic images or those with a different meaning from the original image. Given their performance and good results, GANs have enabled the generation of new images from class labels \cite{mirza_conditional_2014,odena_conditional_2017} and the synthesis of text descriptions \cite{reed_generative_2016,xu_attngan_2018,zhang_stackgan_2017,zhang_stackgan_2019}, which allows the generation of completely new artworks that represent feelings and emotions indicated from text or as classes when training the model.

\subsection{Art Generation using GANs}
The emergence of GANs had a significant impact on the generation of artistic works, whether transforming photographic images into paintings or generating completely new works. Nakano et al. \cite{nakano_neural_2019} present ''Neural Painters'', a generative model of brush strokes learned from a real, non-differentiable, and non-deterministic program. They propose a method to ''motivate'' an agent to paint using more human-like brush strokes when reconstructing digits. \citet{huang_learning_2019} presents a method to teach machines to paint like humans, who are capable of using small brush strokes to achieve excellent results in their paintings. The goal of their model is to decompose the original image into different brush strokes and then recreate them on the canvas. To mimic the human painting process, the agent is trained to predict the next brush stroke based on the current state of the canvas and the reference image to be painted.

A challenge that has been worked on in recent years is the generation of new artistic images or images with a different meaning from the original. The emergence of GANs \cite{goodfellow_generative_2014} and the popularity they have gained in recent years, given their performance and good results, definitely show potential to achieve this goal. GANs have allowed the generation of new images from class labels \cite{mirza_conditional_2014, odena_conditional_2017} and by synthesizing text descriptions \cite{reed_generative_2016, xu_attngan_2017,zhang_stackgan_2017,zhang_stackgan_2018}, which would allow the generation of completely new artistic works that represent feelings and emotions indicated in the form of text or as classes when training new models. \citet{zhang_ai_2018} present an approach for generating artistic works with a specific artistic genre, based on the content text given by the user. They build an input and output system called ''AI Painting'', which consists of three parts: the content, which is an object or scene written in natural language; a word for aesthetic effect, for example, cheerful or depressive; and an artistic genre, for example, impressionism or suprematism. The workflow of this method consists of four steps: 1) generate an image based on the natural language content input using StackGAN++ \cite{zhang_stackgan_2018}; 2) modify the image to include the specified aesthetic effect; 3) transfer the image to the corresponding genre using neural style transfer; 4) illustrate the painting process in a short video.

\citet{li_abstract_2020} present a method for generating abstract paintings. Using the WikiArt dataset and a k-means algorithm that automatically finds the optimal value of k for color segmentation, they manage to divide each painting into color blocks. Then the image segmented by color blocks is used as a real input image to the discriminator, teaching the generator to paint abstract images with color blocks.

\citet{lisi_modelling_2020} introduce a new cGAN training method, which allows the generation of samples from a sequence of distributions. The training was carried out with paintings from a series of artistic movements, which represented a different distribution. Discoveries in each distribution can be used by cGANs to predict ''future'' paintings. The experiments demonstrate that this training is capable of generating accurate predictions of future art, using paintings from the past as a training dataset.

\citet{ozgen_words_2020} investigated the generation of artistic works in a varied dataset, which includes images with variations in color, shapes, and content. This variation present in the dataset provides originality, which is very important for artistic creation and its essence. One of the main characteristics of this model is that, instead of using phrases as descriptive input, keywords are used. They propose a sequential architecture of GANs, which first processes the given description and creates a base image, while the following stages focus on creating high-resolution artistic-style images without worrying about working with word vectors.

As can be observed, numerous proposals have emerged over the years aimed at generating artistic works using computational techniques. This ongoing research has led to remarkable results, including paintings that are often indistinguishable from real works. However, while some approaches consider emotions in the generation of artistic works \cite{zhang_ai_2018, bossett_emotion-based_2021}, there are relatively few studies that focus primarily on emotions as the central aspect of the generation process.

\section{Materials and Methods}
\begin{figure*}[ht]
\centering
\includegraphics[width=1.0\textwidth]{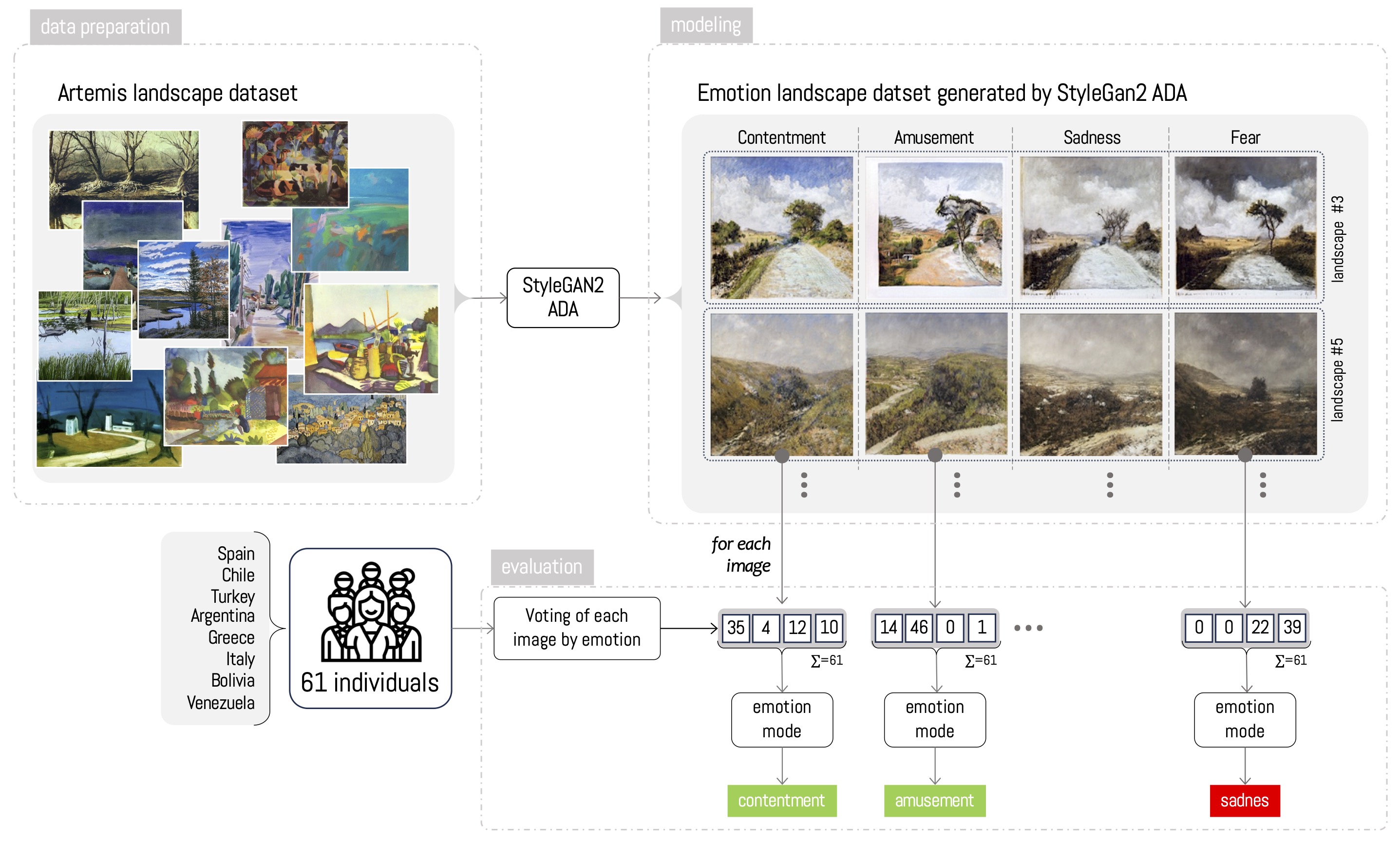}    
\caption{General scheme of the evaluation process of emotions generated by a generative neural. The method comprises three stages: data preparation, modelling and evaluation.}\label{fig_1}
\end{figure*}

\begin{figure*}[ht]
\centering
\includegraphics[width=1.0\textwidth]{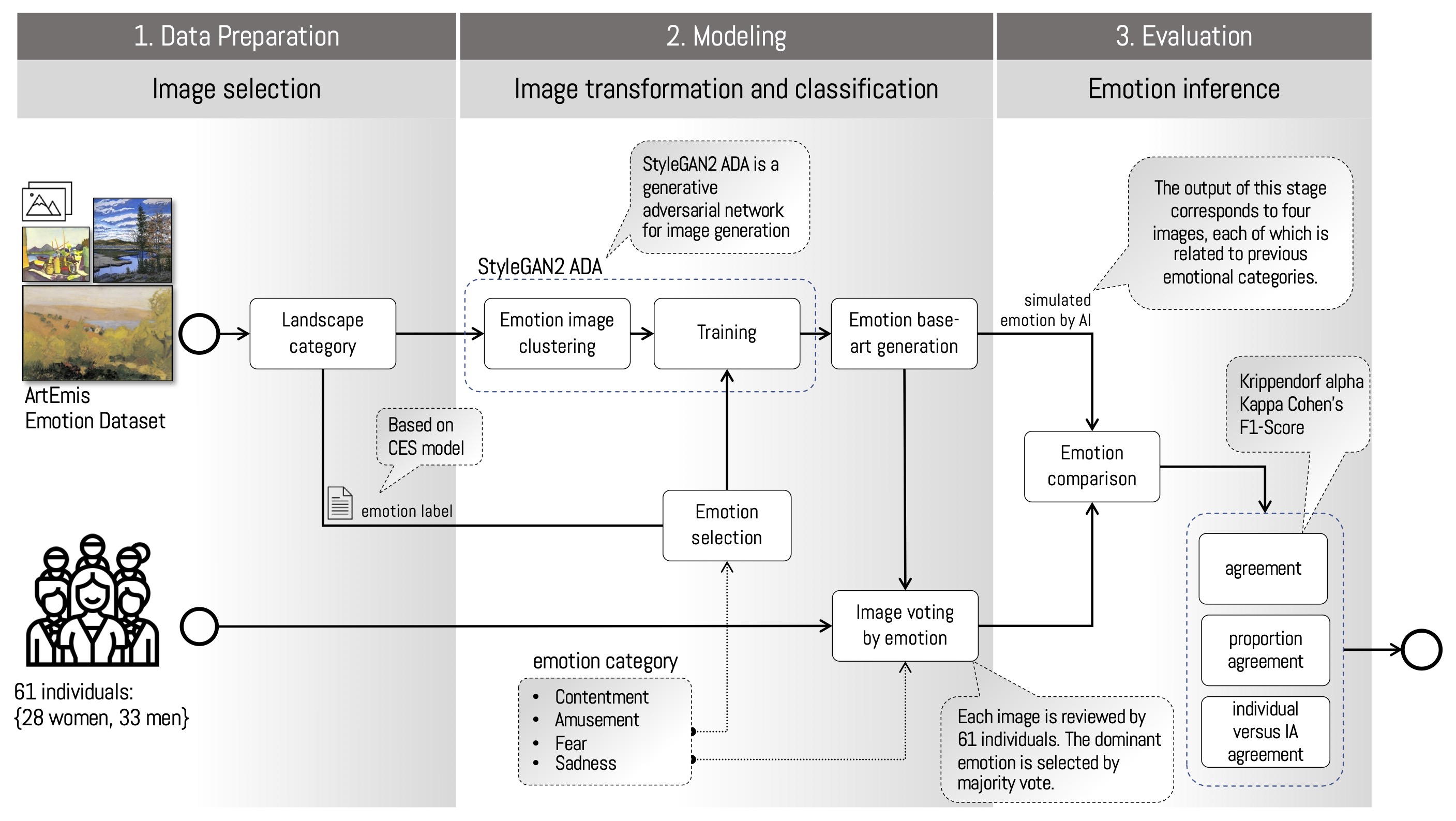}
\caption{Proposed methodology for emotion evaluation generated by a generative network. Within each stage, there are multiple sub-stages dedicated to image development and evaluation.}\label{fig_2}
\end{figure*}

The research was divided into three stages: (1) data preparation, (2) modelling, and (3) evaluation. To provide an overview, we explain each of them in the in-depth process, which is reflected in Fig.\ref{fig_1}. The first phase consisted of data preparation. This process begins with the selection of artworks associated with the landscape category. It is important to note that owing to the type of training of the algorithm, each artwork must be associated with one or more emotions according to an emotional model (which in this case is discrete; \cite{lang_international_2020}. This allows the generative art algorithm to be trained using a specific output class. Second, the modelling phase consisted of 20 landscapes generated by StyleGAN2-ADA tool. Each of these images was associated with one of the four predefined emotions during the training process, corresponding to contentment, amusement, sadness, and fear. In total, 80 images (20 landscape versions of their four emotions) were generated, which were individually evaluated by 61 individuals (33 males, 28 females). Each participant classified each image into one of four emotional categories. The evaluation is blind; that is, the evaluators do not know the emotional category generated by the computer in advance, thus ensuring the independence of the experiment between the evaluator and the generative computational tool. Next, we present each stage in detail at a specific level and the intermediate steps associated with each stage (see Fig. \ref{fig_2}).

\subsection{Data Preparation}
This process involves image extraction and selection with certain emotions and categories. Artemis dataset \cite{achlioptas_artemis_2021} was used, which is composed of 80,031 records obtained from the WikiArt dataset \cite{Mohammad2018WikiArtEA}. Artemis has five records for each artwork: (1) artistic style, (2) artwork, (3) emotion declared by the annotator, (4) explanation by the annotator, and (5) number of annotators who participated in that work. Each record had at least five annotators (evaluators) who defined nine types of emotions per image, corresponding to anger, disgust, fear, and sadness as negative emotions and amusement, awe, contentment, and excitement as positive emotions. In total, Artemis collected 454,684 explanatory statements and emotional responses. 
Although Artemis has ten categories of artistic styles, we used only the landscape category to reduce the level of figuration and identify stimuli and contextual information \cite{dubal_psychophysical_2014,zhang_analyzing_2011} in the identification of emotions for an observer. Thus, 13,358 records in this category were included.

The emotional categories of Artemis are discrete \cite{lang_international_2020}, which means that it is possible to determine the predominant emotion, defined as the one with the highest frequency of votes for a given record. However, some artworks do not have a predominant emotion. Therefore, some records were discarded for our analysis. This occurs when there are few evaluations for a given work, and/or several of them have the same frequency (the same number of votes). Thus, the dataset was reduced to 9,750 valid records in this study.

Finally, because all the records in Artemis have the name of the work, we used this information to download the images in RGB format from the WikiArt dataset using a web scraping technique \cite{hassan_open_2018}. The next process of training the StyleGAN2 ADA neural network was performed using the set of images.

\subsection{Modelling}
As we have previously discussed, despite the existence of different style transfer tools \cite{cai_image_2023}, we have selected StyleGAN2-ADA since related research indicates that this tool generates good results with a reduced amount of training data \cite{karras_analyzing_2020}. This tool has been configured to generate landscape images of the following four emotions: contentment, amusement, fear, and sadness. According to these categories, it is possible to group into positive emotions (contentment and amusement) and negative emotions (fear, sadness). The emotions that have been discarded are astonishment, excitement, anger, and disgust. In the case of astonishment, this can be seen positively and negatively, which would produce a certain ambiguity in its perception \cite{hua_identifying_2016}. The other discarded emotions were excitement, anger, and disgust since they could be subordinate to the selected set, therefore they presented within one of the quadrants of a continuous emotional model \cite{russell_distinguishing_1974}. For this reason, we have finally considered the four emotional categories described above (contentment, amusement, fear, sadness). Furthermore, from the point of view of the continuous emotional scale (CES) \cite{Mohammad2018WikiArtEA} we observe that the selected emotional categories are situated in each of the quadrants of valence and arousal, thus facilitating their differentiation from other similar emotions.

To carry out the training process, we have grouped the images into the four emotional categories described above. The selected images have been pre-processed by reducing their size to 256x256 pixels to be compatible with the training of the neural network. This tool has been configured on a virtual machine with an NVIDIA Tesla T4 graphics card and the Linux operating system, using the Ubuntu 18.04 LTS distribution. Within the operating system, the StyleGAN2-ADA-Pytorch GitHub repository has been cloned and a Python 3.8 virtual environment has been created. The training process was completed in 4 days 6 hours and 58 minutes. In this way, the generative art tool generated 20 landscape images, with their four emotional variants (contentment, amusement, sadness, and fear), thus achieving a total set of 80 images (see examples in Fig.3).

\begin{figure*}[ht]
\centering
\includegraphics[width=1.0\textwidth]{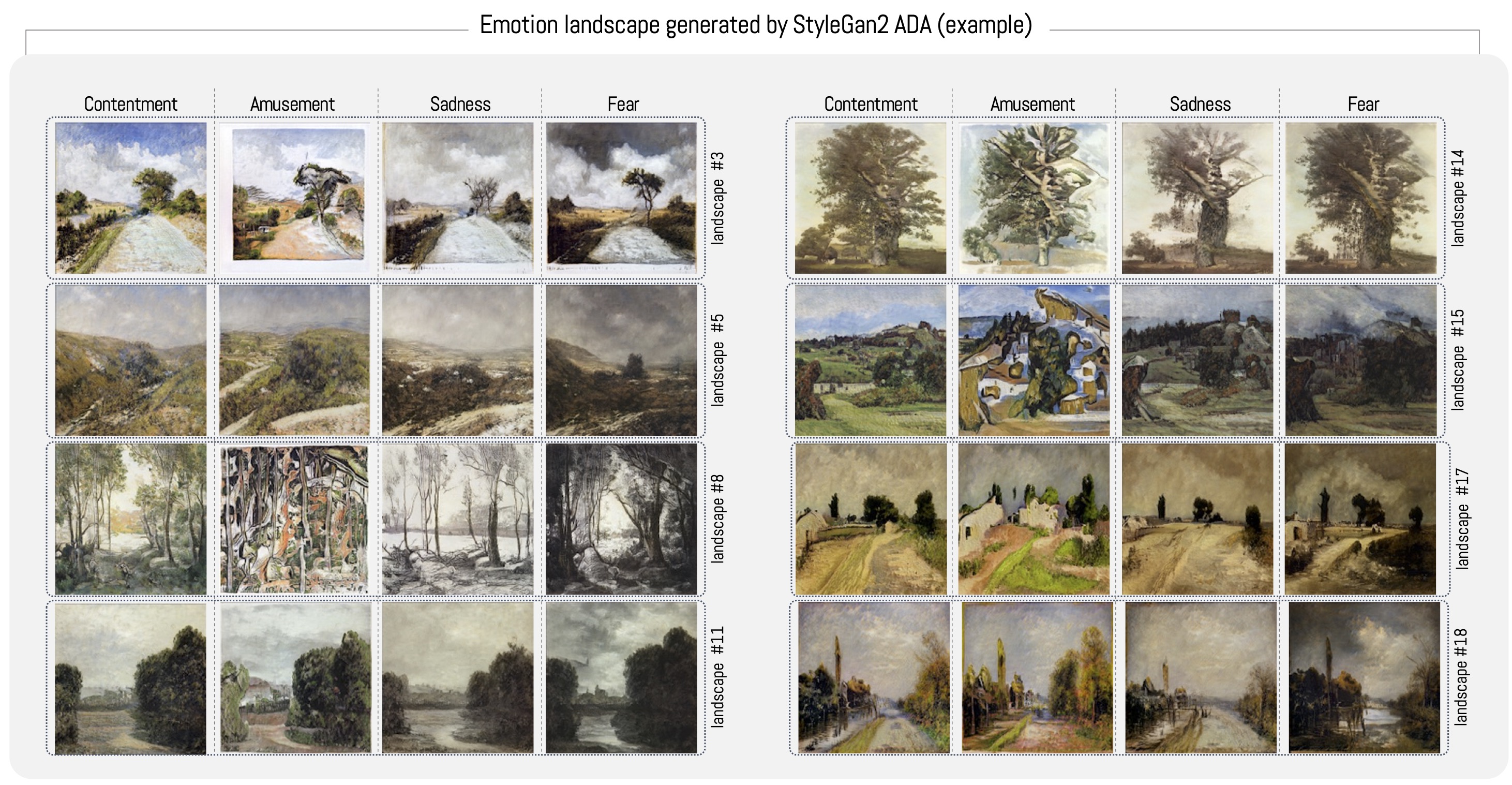}
\caption{Examples of artistic works generated by the StyleGAN2 ADA tool are based on a landscape dataset with four emotional categories. All images are completely new, and there are no existing similar ones in the training set.}\label{fig_3}
\end{figure*}

\subsection{Image voting by emotion}
\begin{figure*}[ht]
\centering
\includegraphics[width=1.0\textwidth]{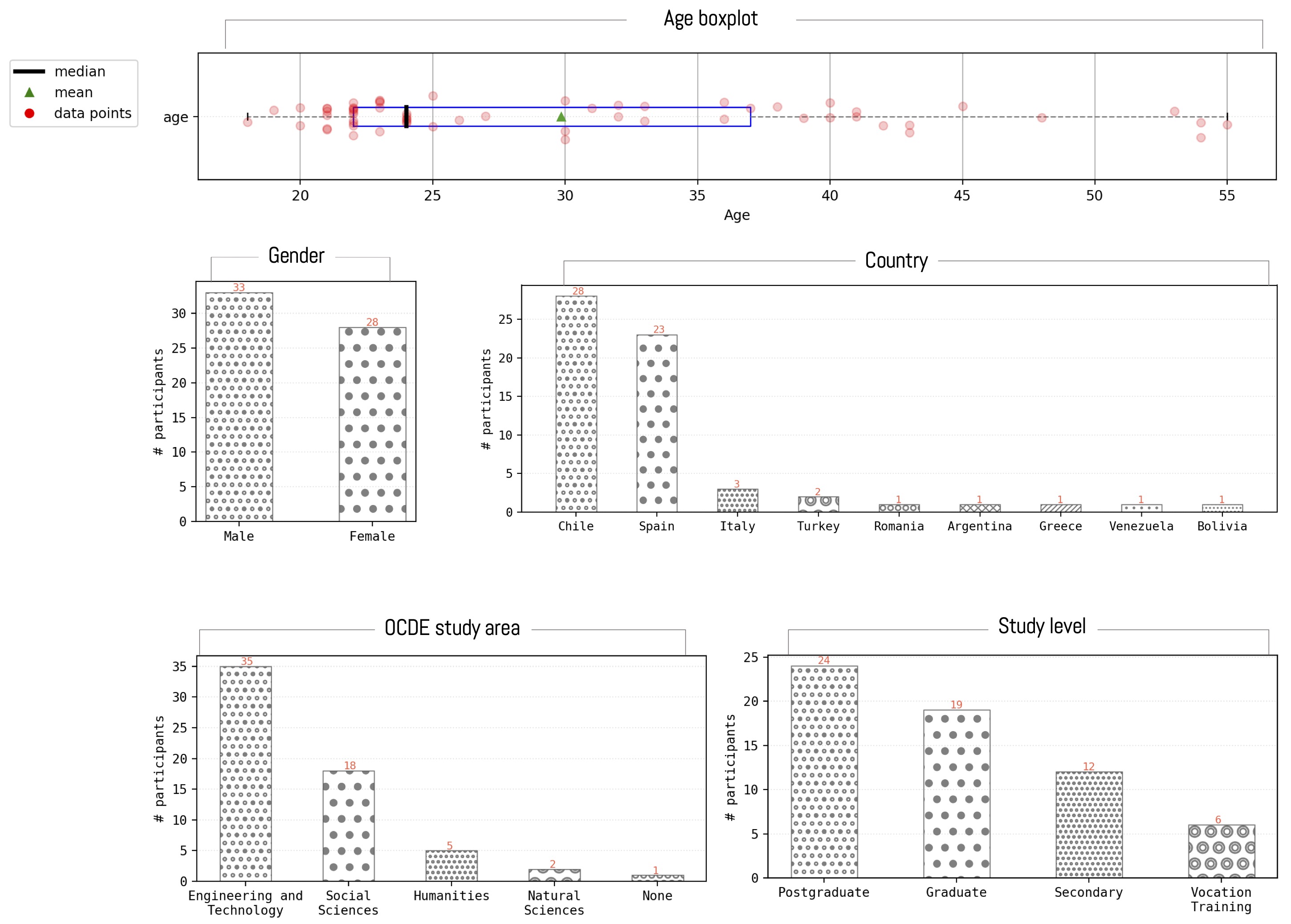}
\caption{Sociodemographic data of study participants: boxplot age, gender {male, female}, country, area of study, highest level of study obtained. More information about the groupings used in the study will be reviewed in the results section.}\label{fig_4}
\end{figure*}

The next step of this research consists of the evaluation of each of the images generated in the previous phase. For this, a form was designed using the Google Form platform, where demographic data was collected regarding age, gender, nationality, level of education and area of knowledge. For the latter, we followed the classification of knowledge area proposed by the Organisation for Economic Co-operation and Development~\cite{oecd_frascati_2015}. In the same form, the automatically generated landscapes were presented in their four emotional versions (80 in total) randomly, so as not to influence the evaluators by any pre-established order. Participants had to indicate one emotion out of the four options (contentment, amusement, fear, sadness) for each version of the landscape. The form was available in English and Spanish from October 30 to November 30, 2023. The average age of the evaluators was 30 years ($std=$7) with a median of 24 years, with a minimum of 18 years, and a maximum of 55 years. Regarding gender, 33 participants declared themselves as male and 28 as female. There were no participants who indicated belonging to another gender (non-binary, or no information). About the area of study, 35\% of the participants declared being associated with the area of engineering and technology, 29\% with the area of social sciences. The areas of humanities and natural sciences together represent only 11\%. Finally, 70\% of the participants declared belonging to the group of graduate or postgraduate as the highest level of education attained. The remaining 30\% are grouped into students who have obtained a professional or high school degree (see indicators in Fig.\ref{fig_4}).

\begin{figure*}[ht]
\centering
\includegraphics[width=1.0\textwidth]{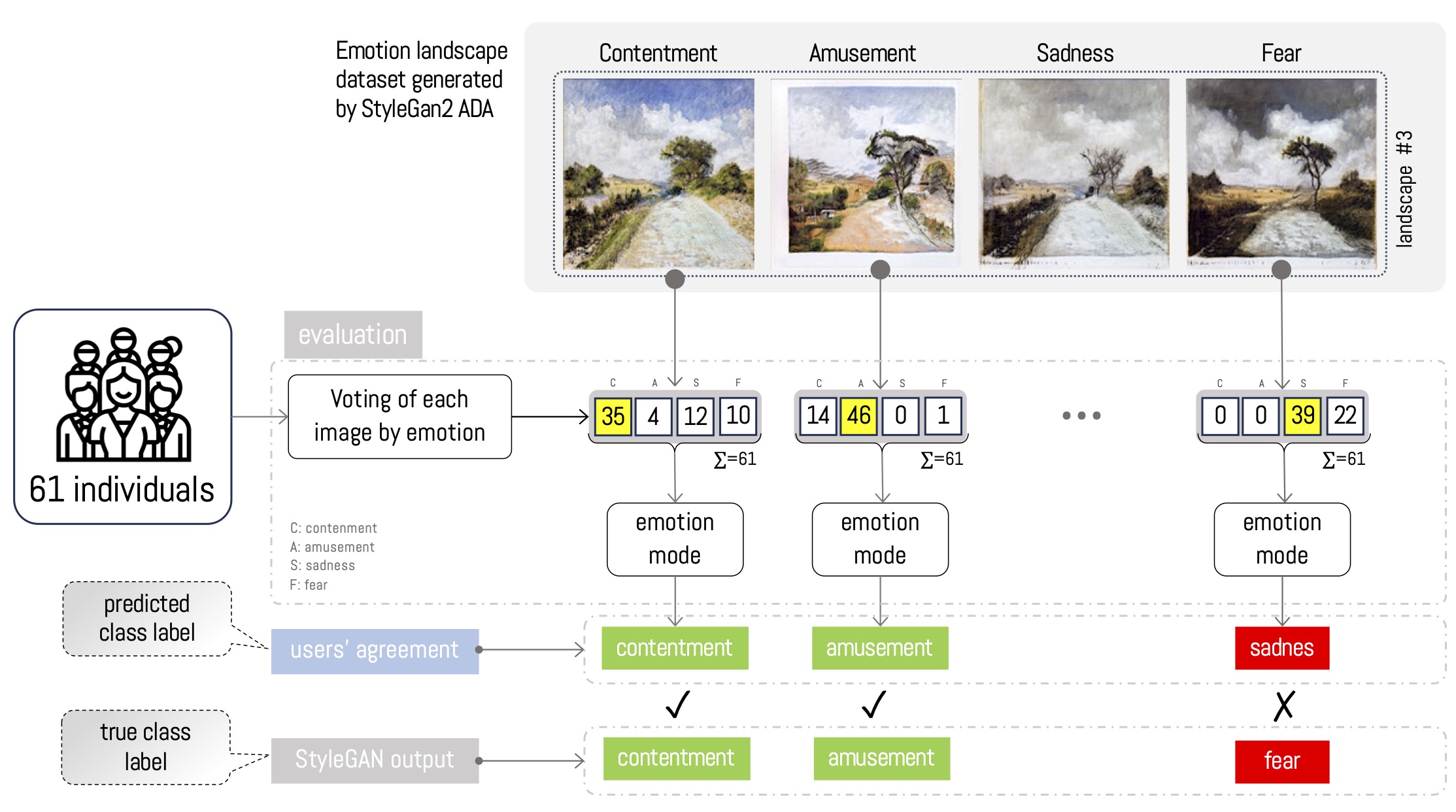}
\caption{Evaluation process and agreement between mode and the StyleGAN2 ADA tool. Each votes on each of the images. Then the mode is calculated for each image to obtain the representative emotion of each image which is compared with the emotional label generated by the generative tool.}\label{fig_5}
\end{figure*}

\subsection{Evaluation}
     Finally, with the data obtained in the previous phase, a statistical analysis has been carried out to evaluate three aspects: the agreement between evaluators, the agreement between the participants (mode) and the generative tool, and a comparative analysis of the agreement reached between different groups of observers and the generative tool. This analysis has been carried out on the 80 images in the four emotional categories (contentment, amusement, fear, sadness). Additionally, the same analysis grouped into positive (contentment and amusement) and negative (fear and sadness) categories.

\subsubsection{Agreement between evaluators}
This process consists of analyzing the inter-rater agreement among the survey participants when emotionally classifying the images generated by the generative tool to measure the agreement between them. For this, we use Krippendorff's Alpha coefficient (\cite{krippendorff_reliability_2004}), which evaluates the level of agreement between observers or participants in assigning categories to a data set. Unlike other indicators, it can be calculated for more than two evaluators, with different types of variables and metrics, in the case of missing data and for small samples \cite{eser_comparison_2022,hayes_answering_2007,volkmann_evaluation_2019}. This step aims to assess whether images produced by the generative tool elicit consistent responses across all participants. This will serve as a proxy to analyze the agreement between each participant and the generative tool itself.

\subsubsection{Agreement between mode and SG2-ADA}
For the evaluation of the agreement between the participants and the generative tool, three aspects are analyzed: the inter-rater agreement, the fisher test and the confusion matrix. In this case, the inter-rater agreement is calculated by taking one of the evaluators as a reference and comparing it with the other observers. Specifically, we take as a reference the label with which the images have been created by the generative tool and compare it with the predominant classification of the participants, that is, with the mode. Therefore, we evaluate the agreement between two evaluators (AI-mode) using Cohen's Kappa coefficient \cite{cohen_coefficient_1960}, following the recommendations to use more than one concordance index in a study \cite{eser_comparison_2022}. Unlike Krippendorff's Alpha coefficient used in the previous section, Cohen's kappa coefficient only allows the analysis between two evaluators, so, in this case, we will use the mode and the emotional label used to generate the images with the generative tool. In this way, it is feasible to determine -per image- the level of agreement or concordance between the evaluators and the generative tool (see an example of the process in Fig. \ref{fig_5}).

On the other hand, we propose in this new methodology the utilization of the confusion matrix, which is habitually used to evaluate the performance of a classification model. The objective of this process is to compare the classification carried out by the participants in the questionnaire of the images into the four emotions with the label assigned by the generative tool. For the construction of the confusion matrix, we define the true class as that class which is generated by the generative tool, and the predicted class as that defined as the mode of the classification of the participants. The precision, recall, and F1 score metrics of the confusion matrix are also calculated to determine the prediction level obtained as if it were a classification problem. With this, we compare the Precision and Recall metrics, obtained from the confusion matrices, for different groups, utilizing Gender (Male-Female), Area of Knowledge (Engineering and Technology - Social Sciences) and Level of Education (Undergraduate-Postgraduate) as segmentation variables through Fisher's Test. We chose to compare these groups as they constitute the majority of respondents, providing a representative sample for analysis. Furthermore, the Jaccard index is utilized, which allows for the determination of the level of intersection between the exposed results among different data sets \cite{costa_further_2021}.

\subsubsection{Proportion analysis}
To evaluate whether the agreement reached in the classification of the images between the participants (mode) and the AI is similar for the images that represent the same emotion, the proportion of agreement is calculated concerning the emotion chosen by the participants (mode) and the emotional label that was provided to the generative tool for each of the 80 images. First, the percentage of agreement is calculated by classifying the images into the four emotional categories, and subsequently, the proportion of agreement is calculated, coding the categories into positive and negative emotions. Then, in both cases, a proportion comparison test is carried out by Cochran's Test, using the emotional label with which the images were generated as the grouping variable.

\section{Results}
The results are presented below according to each of the evaluation stages described in the methodology (Fig.\ref{fig_2}). In particular, we address the results of agreement, agreement between the IA and individuals and the proportion of agreement.

\subsection{Agreement between evaluators}
At the level of comparison on the classifications made by the study participants, the results indicate that people do not agree with each other when classifying the images into four emotional categories (contentment, amusement, fear, and sadness). However, when the emotions are dichotomized into positive and negative, the indicator slightly increases according to Krippendorff's Alpha (see results in Table~ \ref{table_1}).

\begin{table*}[ht]
\caption{Level of agreement according to segmentation by group and emotional category}\label{table_1}%
\begin{tabular}{llcc}
               &    & \multicolumn{1}{l}{4 category}        & \multicolumn{1}{l}{2 category}        \\
               & $n$  & \multicolumn{1}{l}{Kripperdorf alpha} & \multicolumn{1}{l}{Kripperdorf alpha} \\
\midrule
All evaluators & 61 & 0.2284                                & 0.452                                 \\
\midrule
Female       & 28 & 0.2326                                & 0.453                                 \\
Male      & 33 & 0.2216                                & 0.454                                 \\
\midrule
Social Science & 18 & 0.2454                                & \footnotemark[1]0.490                                  \\
Engineering and Technology    & 35 & 0.2195                                & 0.451                                 \\
\midrule
Spain          & 23 & 0.2268                                & 0.437                                \\
Chile          & 28 & 0.2515                                & \footnotemark[1]0.472                               \\
\end{tabular}
\footnotetext[1]{level of agreement is significant at 5\%.}
\end{table*}

When analyzing the agreement according to some group segmentation (gender, country, or area of study), we observe slight differences between the different coefficients. The most relevant level of agreement occurs in the group of the social science knowledge area with the dichotomous emotions (0.4900), followed by the grouping by nationality (Chile 0.4721) (see Table~\ref{table_1}).

\subsection{Agreement between mode and generative IA}
This section analyzes the agreement between participant responses and the output of the generative tool (StyleGAN2-ADA). For this, we use the mode of the evaluators' classifications and the emotion used to generate the images. 

Assuming that the emotion generated by the generative tool corresponds to the actual (or true) class, we analyze the precision, recall and F1 score of the obtained data to quantify the level of agreement in the classifications for each group of evaluators and the generative tool. The results reveal important differences according to the group and the emotional category. As stated in Table~\ref{table_2}, the best classification results were obtained for the Fear category in most groups, however, the performance changes according to the group analyzed. For example, in the female group, an F1-score=0.76 was obtained, and in the same emotional category, we achieved an F1-score=0.89 for the postgraduate group. The above indicates that by maintaining the same emotional category, different groups of segmentation obtain different performances. In the opposite direction, we observe that for the emotional category contentment, there is a lower level of classification for all groups analyzed. The above could indicate that for individuals it is more complex to classify a positive emotion over a negative one. On the other hand, when the emotions are binarized into positive and negative categories, we obtain a high performance in general. However, in some cases, it is observed that the detection of positive emotions is more difficult than negative emotions (see Table \ref{table_3}). To further analyze the statistical differences, we analyze this point in the following section. 

\begin{table*}[ht]
\caption{Level of agreement according to segmentation by group and emotional category}\label{table_2}%
\begin{tabular}{lrrrrrr}
Genre & \multicolumn{3}{c}{Female $n=28$} & \multicolumn{3}{c}{Male $n=33$} \\ \hline
\textbf{Emotion} & \multicolumn{1}{c}{Precision} & \multicolumn{1}{c}{Recall} & \multicolumn{1}{c}{F1-Score} & \multicolumn{1}{c}{Precision} & \multicolumn{1}{c}{Recall} & \multicolumn{1}{c}{F1-Score} \\
Contentment & 0.58 & 0.7 & 0.64 & 0.67 & 0.8 & 0.73 \\
Amusement & 0.81 & 0.65 & 0.72 & 0.94 & 0.75 & 0.83 \\
\textbf{Fear} & \textbf{0.82} & \textbf{0.7} & \textbf{0.76} & \textbf{0.89} & \textbf{0.85} & \textbf{0.87} \\
Sadness & 0.65 & 0.75 & 0.7 & 0.76 & 0.8 & 0.78 \\
 & \multicolumn{1}{l}{} & \multicolumn{1}{l}{} & \multicolumn{1}{l}{} & \multicolumn{1}{l}{} & \multicolumn{1}{l}{} & \multicolumn{1}{l}{} \\
 OCDE study area & \multicolumn{3}{c}{Social Science n=17} & \multicolumn{3}{c}{Engineering and Tech. n=34} \\ \hline
\textbf{Emotion} & \multicolumn{1}{l}{Precision} & \multicolumn{1}{l}{Recall} & \multicolumn{1}{l}{F1-Score} & \multicolumn{1}{l}{Precision} & \multicolumn{1}{l}{Recall} & \multicolumn{1}{l}{F1-Score} \\
Contentment & 0.56 & 0.7 & 0.62 & 0.67 & 0.8 & 0.73 \\
Amusement & \textbf{0.76} & \textbf{0.65} & \textbf{0.7} & 0.88 & 0.75 & 0.81 \\
\textbf{Fear} & \textbf{0.76} & \textbf{0.65} & \textbf{0.7} & \textbf{0.86} & \textbf{0.9} & \textbf{0.88} \\
Sadness & 0.62 & 0.65 & 0.63 & 0.89 & 0.8 & 0.84 \\
 & \multicolumn{1}{l}{} & \multicolumn{1}{l}{} & \multicolumn{1}{l}{} & \multicolumn{1}{l}{} & \multicolumn{1}{l}{} & \multicolumn{1}{l}{} \\
Country & \multicolumn{3}{c}{Spain n= 23} & \multicolumn{3}{c}{Chile n=28} \\ \hline
\textbf{Emotion} & \multicolumn{1}{l}{Precision} & \multicolumn{1}{l}{Recall} & \multicolumn{1}{l}{F1-Score} & \multicolumn{1}{l}{Precision} & \multicolumn{1}{l}{Recall} & \multicolumn{1}{l}{F1-Score} \\
Contentment & 0.57 & 0.65 & 0.6 & 0.71 & 0.85 & 0.77 \\
Amusement & 0.71 & 0.6 & 0.65 & 0.94 & 0.8 & 0.86 \\
\textbf{Fear} & \textbf{0.88} & \textbf{0.7} & \textbf{0.78} & \textbf{0.95} & \textbf{0.9} & \textbf{0.92} \\
Sadness & 0.62 & 0.75 & 0.68 & 0.8 & 0.8 & 0.8 \\
 & \multicolumn{1}{l}{} & \multicolumn{1}{l}{} & \multicolumn{1}{l}{} & \multicolumn{1}{l}{} & \multicolumn{1}{l}{} & \multicolumn{1}{l}{} \\
 Study level & \multicolumn{3}{c}{Postgraduate} & \multicolumn{3}{c}{Graduate} \\ \hline
\textbf{Emotion} & \multicolumn{1}{l}{Precision} & \multicolumn{1}{l}{Recall} & \multicolumn{1}{l}{F1-Score} & \multicolumn{1}{l}{Precision} & \multicolumn{1}{l}{Recall} & \multicolumn{1}{l}{F1-Score} \\
Contentment & 0.54 & 0.70 & 0.61 & 0.62 & 0.75 & 0.68 \\
Amusement & 0.71 & 0.50 & 0.59 & 0.75 & 0.75 & 0.75 \\
\textbf{Fear} & \textbf{0.94} & \textbf{0.85} & \textbf{0.89} & \textbf{0.84} & \textbf{0.8} & \textbf{0.82} \\
Sadness & 0.73 & 0.80 & 0.76 & 0.82 & 0.7 & 0.76  \\ \hline
\end{tabular}
\end{table*}

\begin{table*}[ht]
\caption{Level of agreement according to segmentation by group and emotional category}\label{table_3}%
\begin{tabular}{lrrrrrr}
 Genre & \multicolumn{3}{c}{Female $n=28$} & \multicolumn{3}{c}{Male $n=33$} \\ \hline
\textbf{Emotion}  & \multicolumn{1}{c}{Precision} & \multicolumn{1}{c}{Recall} & \multicolumn{1}{c}{F1-Score} & \multicolumn{1}{c}{Precision} & \multicolumn{1}{c}{Recall} & \multicolumn{1}{c}{F1-Score} \\
Positive & 0.91 & 0.97 & 0.94 & 0.95 & 0.95 & 0.95 \\
Negative & 0.97 & 0.9 & 0.94 & 0.95 & 0.95 & 0.95 \\
 & \multicolumn{1}{l}{} & \multicolumn{1}{l}{} & \multicolumn{1}{l}{} & \multicolumn{1}{l}{} & \multicolumn{1}{l}{} & \multicolumn{1}{l}{} \\
 OCDE study area & \multicolumn{3}{c}{Social Science n=17} & \multicolumn{3}{c}{Engineering and Tech. n=34} \\ \hline
\textbf{Emotion} & \multicolumn{1}{l}{Precision} & \multicolumn{1}{l}{Recall} & \multicolumn{1}{l}{F1-Score} & \multicolumn{1}{l}{Precision} & \multicolumn{1}{l}{Recall} & \multicolumn{1}{l}{F1-Score} \\
Positive & 0.91 & 0.97 & 0.94 & 0.95 & 0.97 & 0.96 \\
Negative & 0.97 & 0.9 & 0.94 & 0.97 & 0.95 & 0.96 \\
 & \multicolumn{1}{l}{} & \multicolumn{1}{l}{} & \multicolumn{1}{l}{} & \multicolumn{1}{l}{} & \multicolumn{1}{l}{} & \multicolumn{1}{l}{} \\
Country & \multicolumn{3}{c}{Spain n= 23} & \multicolumn{3}{c}{Chile n=28} \\ \hline
\textbf{Emotion} & \multicolumn{1}{l}{Precision} & \multicolumn{1}{l}{Recall} & \multicolumn{1}{l}{F1-Score} & \multicolumn{1}{l}{Precision} & \multicolumn{1}{l}{Recall} & \multicolumn{1}{l}{F1-Score} \\
Positive & 0.89 & 0.97 & 0.93 & 0.93 & 0.93 & 0.93 \\
Negative & 0.97 & 0.88 & 0.92 & 0.93 & 0.93 & 0.93 \\
 & \multicolumn{1}{l}{} & \multicolumn{1}{l}{} & \multicolumn{1}{l}{} & \multicolumn{1}{l}{} & \multicolumn{1}{l}{} & \multicolumn{1}{l}{} \\
 Study level & \multicolumn{3}{c}{Postgraduate} & \multicolumn{3}{c}{Graduate} \\ \hline
\textbf{Emotion} & \multicolumn{1}{l}{Precision} & \multicolumn{1}{l}{Recall} & \multicolumn{1}{l}{F1-Score} & \multicolumn{1}{l}{Precision} & \multicolumn{1}{l}{Recall} & \multicolumn{1}{l}{F1-Score} \\
Positive & 0.95 & 0.93 & 0.94 & 0.95 & 0.97 & 0.96 \\
Negative & 0.93 & 0.95 & 0.94 & 0.97 & 0.95 & 0.96 \\ \hline
\end{tabular}
\end{table*}

We analyze whether there are significant differences in the agreement between the participants (mode) and the AI when classifying the images (in four categories and two categories) between participants groups: men-women, engineering-social sciences, and graduate-postgraduates. To do this, we compare the precision and recall of the confusion matrices (Table~\ref{table_2} and Table \ref{table_3}) using Fisher's Test. When the precision of the confusion matrices is compared, the results show that there are only significant differences in the case of the 'Sadness' emotion. Specifically, precision is higher for men ($p$-value$=$0.007), for individuals in the 'Engineering and Technology' area of knowledge at a 10\% significance level ($p$-value$=$0.07), and for graduates ($p$-value$=$0.042) at 5\% (see Table~\ref{table_8}), showing that these groups coincide to a greater extent with the AI.  

\begin{table*}[ht]
\caption{Comparison of the accuracy of the 4 emotions between different groups (Fisher's test).}\label{table_8}
\begin{tabular}{lccc}
 & \multicolumn{3}{c}{\textit{\textbf{$p$-value\footnotemark[1]}}} \\
\textbf{Emotion} & \textbf{Male-Female} & \textbf{Eng.Tech. - Social Sciences} & \textbf{Graduate-Postgraduate} \\ \hline
Contentment & 0.383 & 0.319 & 0.37 \\
Amusement & 0.3 & 0.328 & 0.56 \\
Fear & 0.445 & 0.492 & 0.323 \\
Sadness & 0.007***↑ & 0.070*↑ & 0.042**↑ \\ \hline
\end{tabular}
\footnotetext[1]{Significant differences are denoted with asterisks (*p\textless{}0.1, ** p\textless{}0.05, *** p\textless{}0.01).}
\footnotetext[2]{Note: In cases of significant differences, an upward arrow (↑) indicates that the accuracy will be higher for the category mentioned first}
\end{table*}

If emotions are dichotomized into positive and negative, and different groups are compared, the results show that there are no significant differences in precision in any case ($p$-value$>$0.1) (see Table \ref{table_9}).

\begin{table*}[ht]
\caption{Comparison of the accuracy of positive/negative emotions between different groups (Fisher's test).}\label{table_9}
\begin{tabular}{llll}
\textbf{} & \multicolumn{3}{c}{\textit{\textbf{$p$-value}}} \\
\textbf{Emotion} & \multicolumn{1}{c}{\textbf{Male-Female}} & \multicolumn{1}{c}{\textbf{Eng.Tech. - Social Sciences}} & \multicolumn{1}{c}{\textbf{Graduate-Postgraduate}} \\ \hline
Negative & 0.683 & 0.74 & 0.327 \\
Positive & 0.512 & 0.361 & 0.673 \\ \hline
\end{tabular}
\end{table*}

As we can observe in Table \ref{table_10}, the comparison of Recall for the different groups shows that there are only statistically significant differences in the classification of images of the 'Fear' emotion between the two categories of the 'Area of Knowledge' variable at 10\%, with Recall being higher for the 'Engineering and Technology' group ($p$-value$=$0.064).

\begin{table*}[ht]
\caption{Comparison of the Recall of the four emotions between different groups (Fisher's test).}\label{table_10}
\begin{tabular}{lccc}
 & \multicolumn{3}{c}{\textit{\textbf{$p$-value}}} \\
\textbf{Emotion} & \textbf{Male-Female} & \textbf{Eng.Tech. - Social Sciences} & \textbf{Graduate-Postgraduate} \\ \hline
Contentment & 0.358 & 0.358 & 0.5 \\
Amusement & 0.366 & 0.366 & 0.1 \\
Fear & 0.255 & 0.064*↑ & 0.5 \\
Sadness & 0.5 & 0.366 & 0.358 \\ \hline
\end{tabular}
\footnotetext[1]{Significant differences are denoted with asterisks (*p\textless{}0.1, ** p\textless{}0.05, *** p\textless{}0.01).}
\footnotetext[2]{Note: in case of significant differences ↑ indicates that the accuracy will be higher for the category that is named first.}
\end{table*}

The results in the case of the dichotomous Emotion variable show that there are no significant differences in Recall in any case ($p$-value$>$0.1) (Table \ref{table_11}).

\begin{table*}[ht]
\caption{Comparison of positive/negative emotion recall between different groups (Fisher's test).}\label{table_11}
\begin{tabular}{lccc}
\textbf{} & \multicolumn{3}{c}{\textit{\textbf{$p$-value}}} \\
Emotion & \textbf{Male-Female} & \textbf{Eng.Tech. - Social Sciences} & \textbf{Graduate-Postgraduate} \\ \hline
Negative & 0.5 & 0.338 & 0.692 \\
Positive & 0.692 & 0.753 & 0.308 \\ \hline
\end{tabular}
\end{table*}

To further analyze the results, we use the Jaccard index, which allows us to evaluate the level of intersection between the participants' responses for the image generated by the generative tool. The results of this indicator point to relevant differences between four emotional categories versus two emotional categories (positive and negative). It is observed that there is a greater intersection between the responses of the participants in two categories compared to four emotional categories. This is because there is greater agreement with the emotion expressed by the generative tool when the emotional categories are positive and negative. On the contrary, when we increase the number of emotions, users do not reflect an agreement with what is expressed by the tool. These results are consistent with those previously obtained in Tables \ref{table_2} and \ref{table_3}. 

A relevant result is shown in the Social Science group, where the highest level of agreement with the generative tool is obtained for two emotions (70.08\%). The same situation occurs for the participants from Chile, reaching 70.18\% agreement with the tool. In general, a 68.49\% agreement is obtained between all participants and the generative tool only when we binarize the emotions. This result drops to 37.06\% when we have four emotional categories (Table~\ref{table_4}).

\begin{figure*}[ht]
\centering
\includegraphics[width=1.0\textwidth]{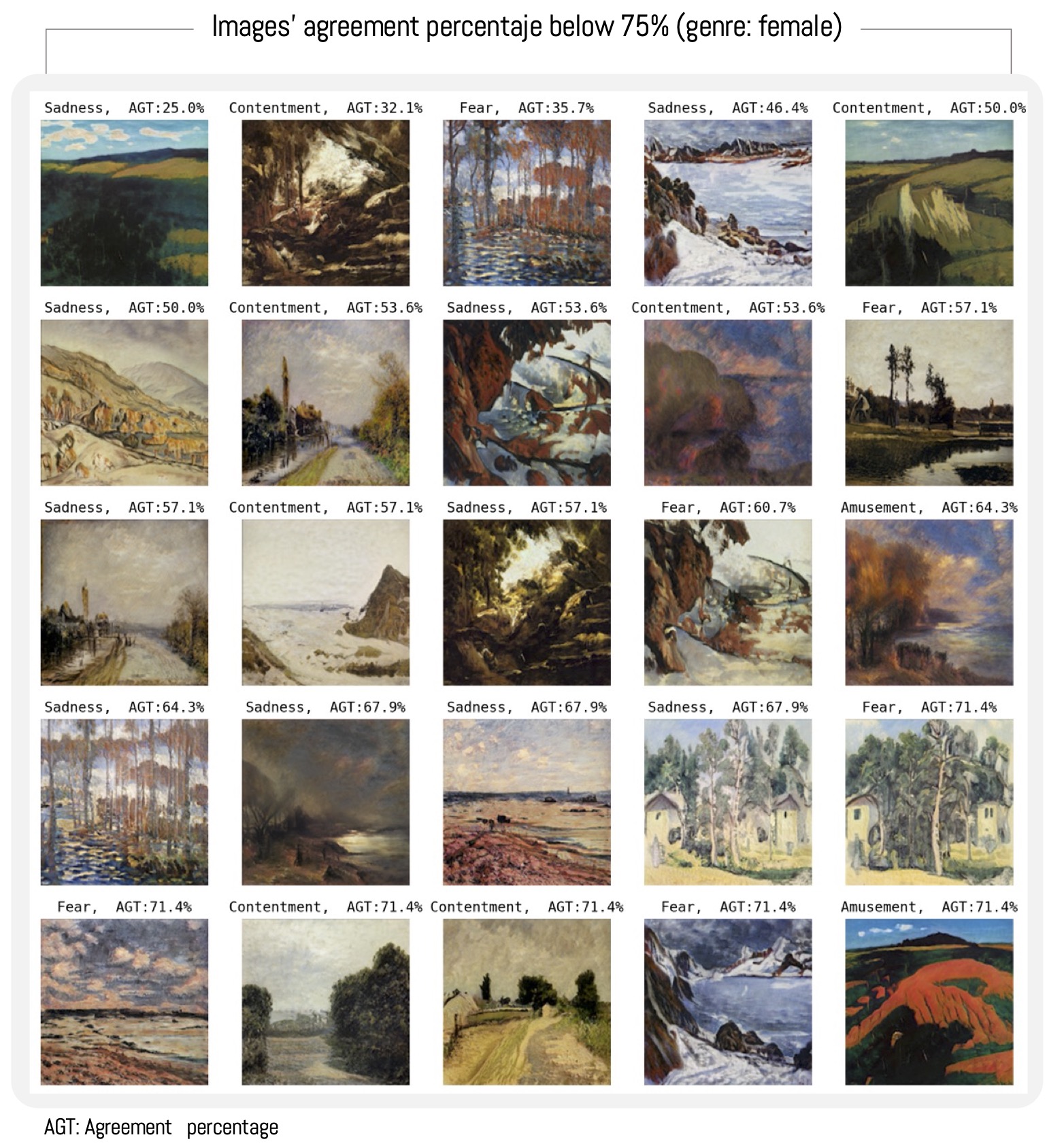}
\caption{Percentage of agreement with Jaccard's index for the female gender with the generative tool under 75\%.}\label{fig_6}
\end{figure*}

\begin{figure*}[ht]
\centering
\includegraphics[width=1.0\textwidth]{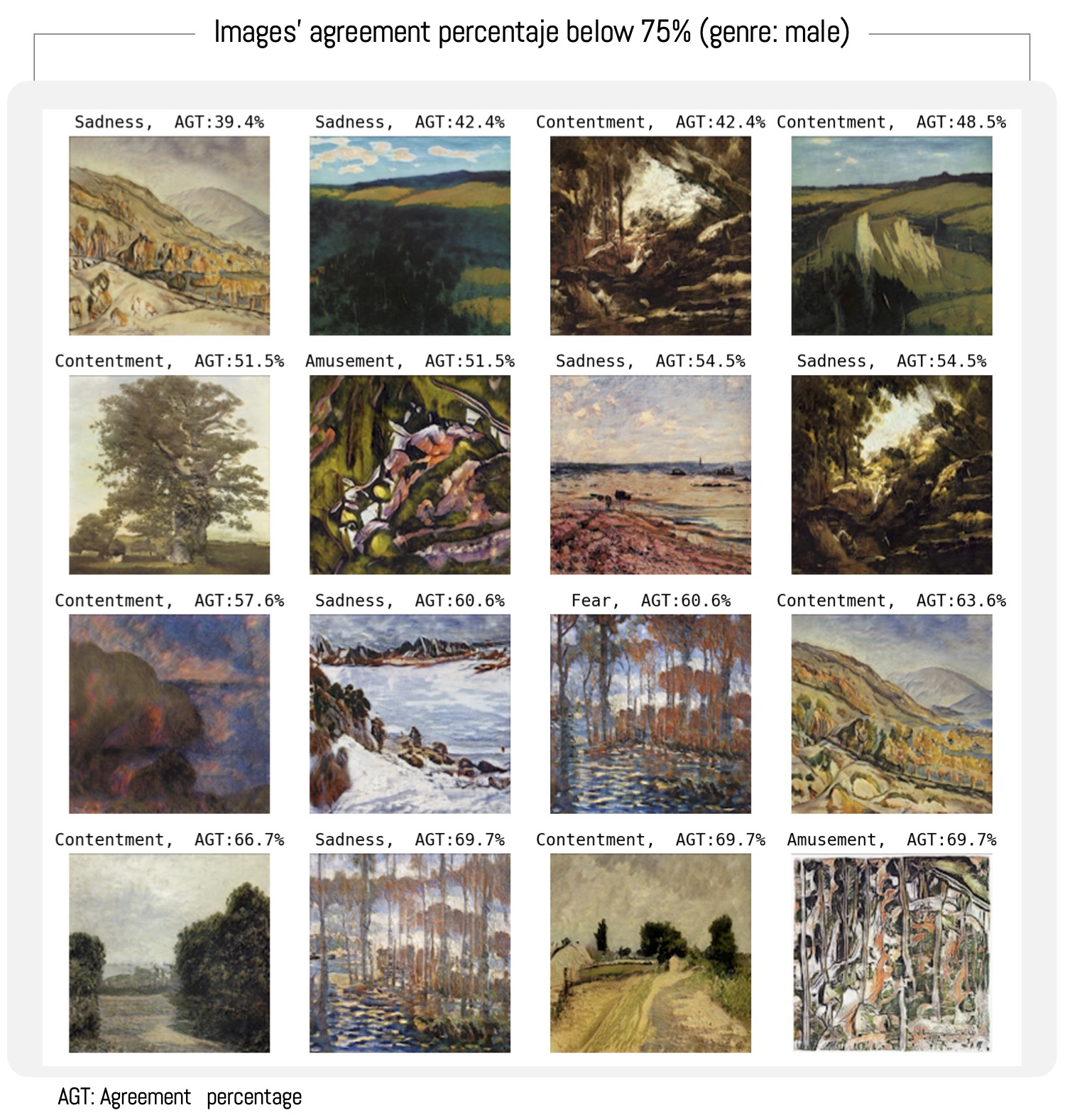}
\caption{Percentage of agreement with Jaccard's index for the male gender with the generative tool under 75\%.}\label{fig_7}
\end{figure*}

To visualize the most relevant results at the intersection level, Figures \ref{fig_6} and \ref{fig_7} illustrate all the images that obtained a percentage lower than 75\% in the Jaccard index with the generative tool. As can be seen in Fig. \ref{fig_6}, there is indiscriminate disagreement in both positive and negative images for both male and female genders, however, the proportion of correct answers is more balanced between positive emotions (contentment and amusement) and negative (fear and sadness).

\begin{table}[ht]
\caption{Jaccard index results segmented by emotional category and analysis group}\label{table_4}%
\begin{tabular}{lccc}
 & \multicolumn{1}{l}{} & 4 category & 2 category \\
 & $n$ & Jaccard index & Jaccard index \\ \hline
All evaluators & 61 & 0.3706 & 0.6849 \\ \hline
Female & 28 & 0.3565 & 0.6735 \\
Male & 33 & 0.3825 & 0.6947 \\ \hline
Social Science & 18 & 0.3645 & 0.7008 \\
Engineering and Technology & 35 & 0.3795 & 0.6939 \\ \hline
Spain & 23 & 0.344 & 0.6587 \\
Chile & 28 & 0.4069 & 0.7018 \\ \hline
\end{tabular}
\end{table}

\begin{table}[ht]
\caption{Results of Cohen's Kappa index segmented by emotional category and group of analysis}\label{table_5}%
\begin{tabular}{lccc}
 & \multicolumn{1}{l}{} & 4 category & 2 category \\
 & $n$ & Cohen's Kappa & Cohen's Kappa \\ \hline
All evaluators & 61 & 0.7 & 0.875 \\ \hline
Female & 28 & 0.616 & 0.85 \\
Male & 33 & 0.716 & 0.85 \\ \hline
Social Science & 18 & 0.5166 & 0.85 \\
Engineering and Technology & 35 & 0.75 & 0.875 \\ \hline
Spain & 23 & 0.55 & 0.8 \\
Chile & 28 & 0.7833 & 0.875 \\ \hline
\end{tabular}
\end{table}

Regarding the classifications with an agreement above 90\% to the emotion generated by the generative tool, differences are observed between the gender categories, although there is one image with the Fear emotion associated, which achieved 100\% accuracy in the classification (Fig. \ref{fig_7}).

To conclude the study in this section, we analyze the level of agreement using Cohen's kappa coefficient (Table \ref{table_5}). Recall that in this process we work with the mode and the emotional label used by the generative tool as evaluators. Following Landis and Koch (\cite{landis_measurement_1977}), the results show an almost perfect agreement using the mode of all observers in the binarized emotional category (k=0.88). This coefficient is repeated in the grouping of Chilean nationality and the area of knowledge in Engineering and Social sciences.

With the four emotional categories, the coefficients are more disparate, reaching in most cases a substantial level of agreement, except in the grouping of Spanish nationality and the area of social science study, which would be within the moderate agreement \cite{landis_measurement_1977}. The highest percentage of agreement is obtained by the Chilean nationality grouping (k=0.78, Table~\ref{table_5}).

\begin{figure*}[ht]
\centering
\includegraphics[width=1.0\textwidth]{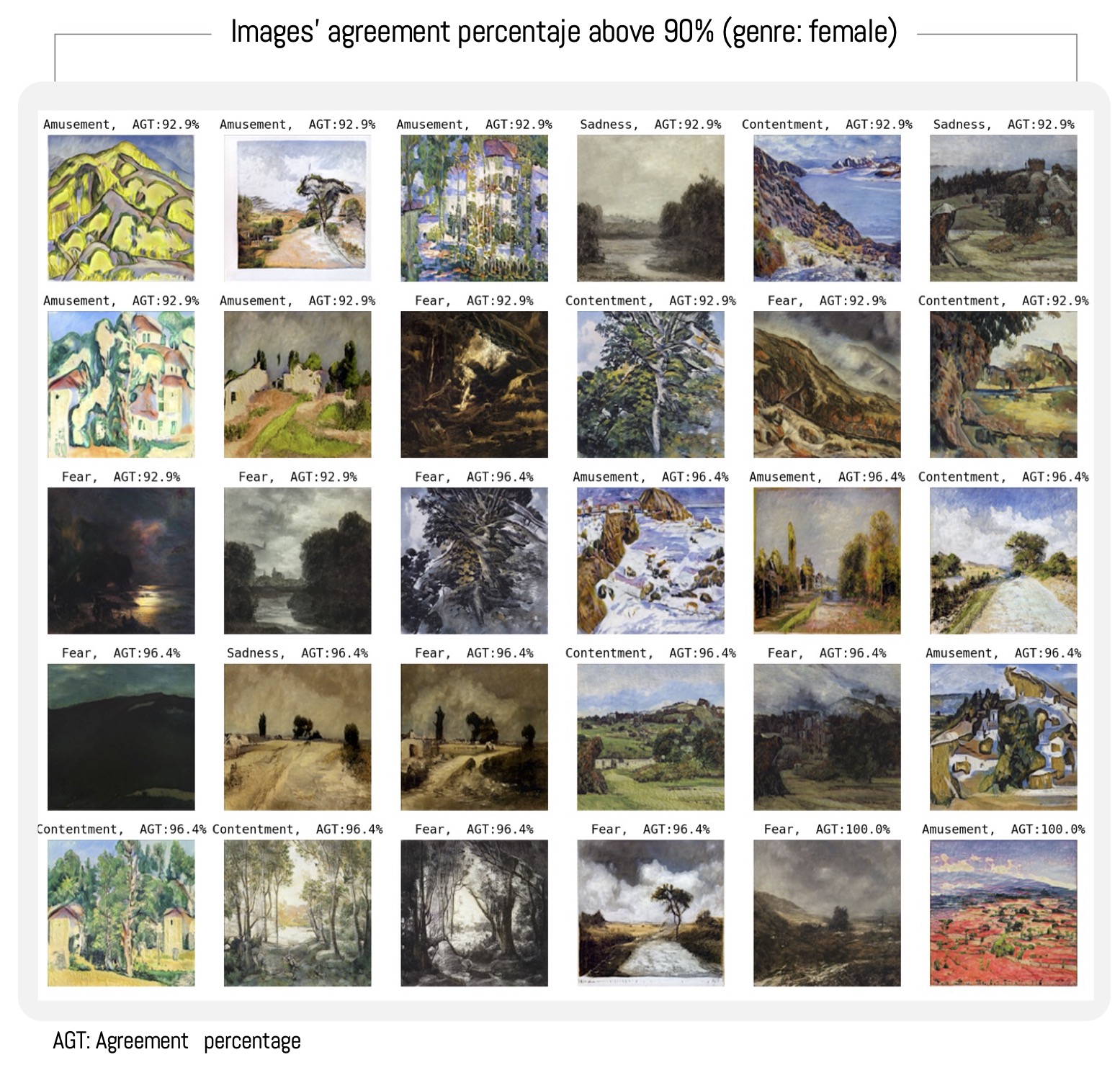}
\caption{Percentage of agreement with Jaccard's index for the female gender with the generative tool over 90\%.}\label{fig_8}
\end{figure*}

\begin{figure*}[ht]
\centering
\includegraphics[width=1.0\textwidth]{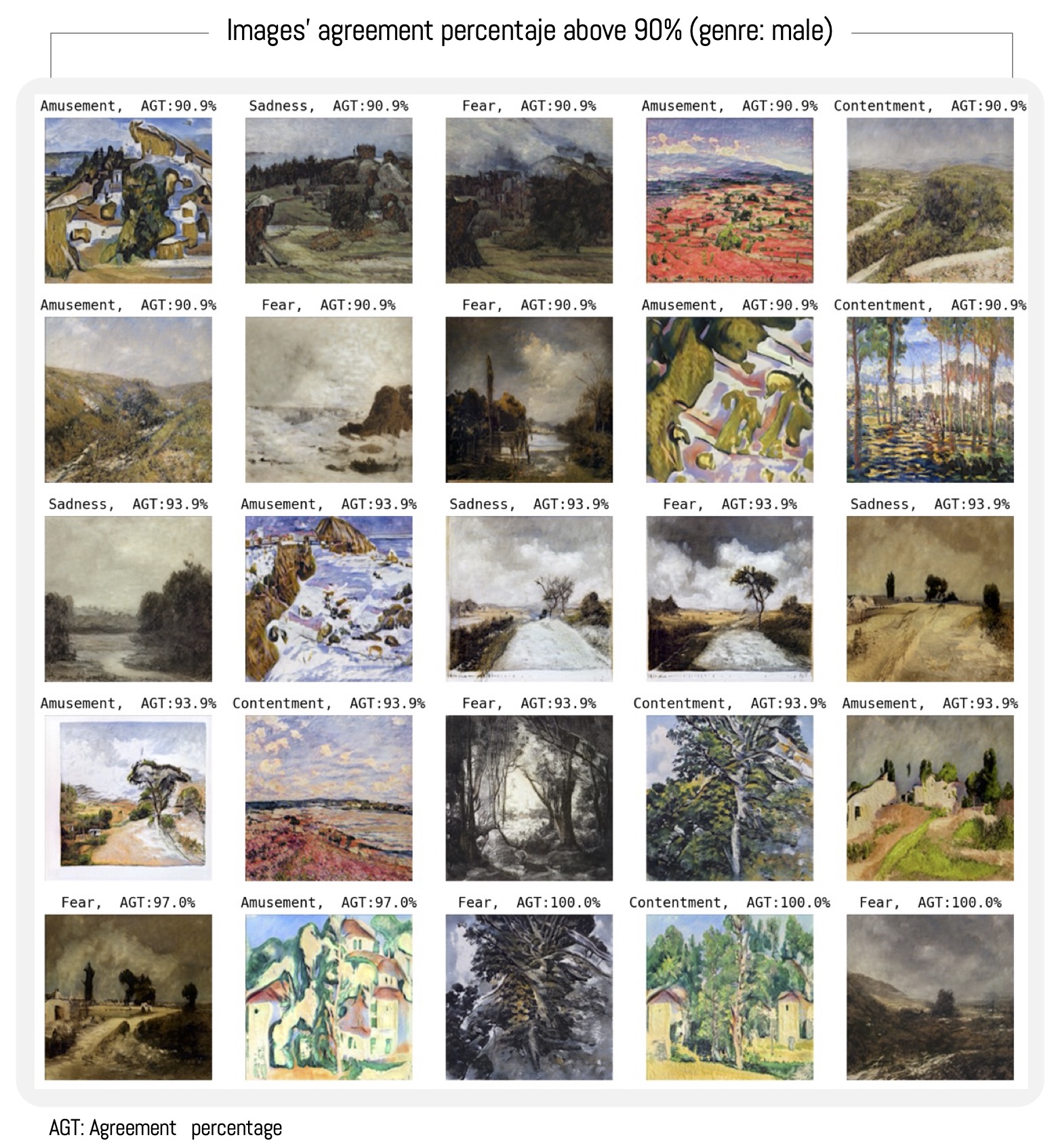}
\caption{Percentage of agreement with Jaccard's index for the male gender with the generative tool over 90\%.}\label{fig_9}
\end{figure*}

\subsection{Analysis of proportions}
This section aims to investigate  whether the proportion of participants agreeing with the generative tool (SD-ADA2 GAN) remains consistent across different images. Specifically, we examine if this proportion remains invariant when individuals categorize the 20 images that share the same ground truth label (generative tool). To achieve this, we first quantify the percentage of participants who agree with the AI's classifications for each of the 80 images. Subsequently, we conduct Cochran's Test to ascertain if statistically significant differences exist among the images generated with the same emotional label.

\begin{figure*}[t!]
\centering
\includegraphics[width=1.0\textwidth]{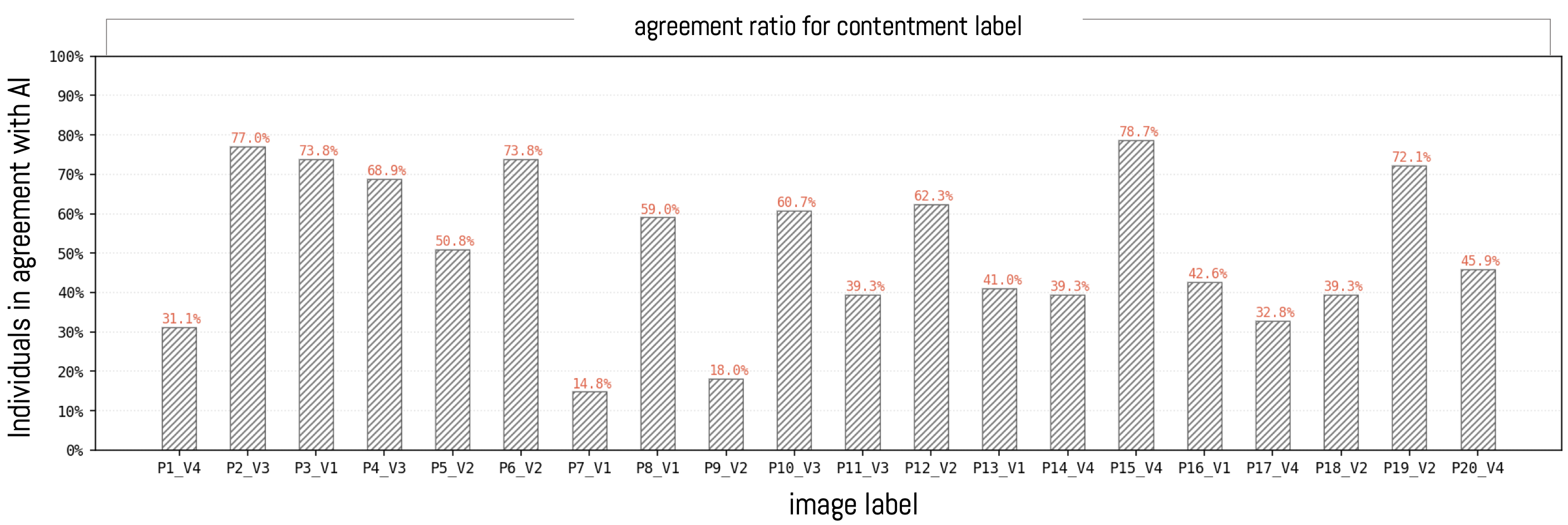}
\caption{Proportion of individuals who agree with the AI when classifying images labelled as Contentment.}\label{fig_10}
\end{figure*}

We represent in Fig.~\ref{fig_10} the percentage of individuals who have agreed with the AI when classifying the image with the emotional label ``contentment''. As we can observe, more than 60\% of the participants have selected the actual label in 10 of the 20 images. There is also diversity in the agreement, with image 15 version 4 being the one in which more individuals agree with the AI, specifically 79\%. In contrast to this percentage, and at the other extreme, only 15\% of participants agree with the AI when classifying image 7 version 1. We can affirm that these differences are statistically significant ($p$-value$<$0.001. See Table \ref{table_6}).

\begin{figure*}[t!]
\centering
\includegraphics[width=1.0\textwidth]{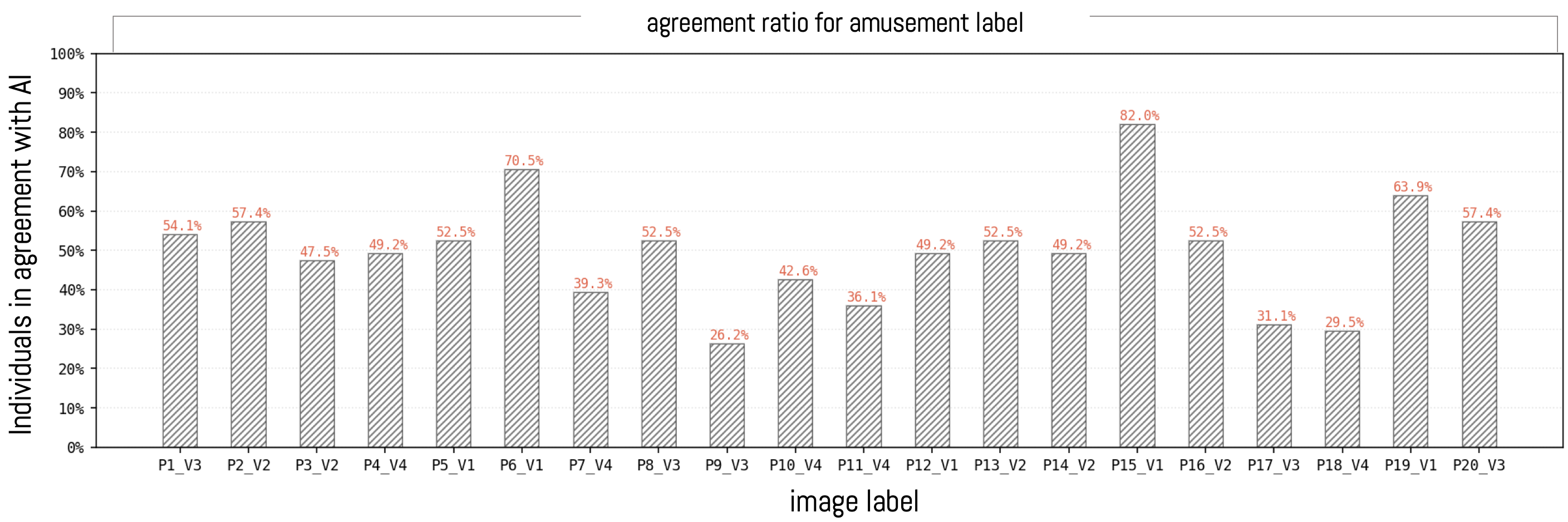}
\caption{Proportion of individuals who agree with the AI when classifying images labelled as Amusement}\label{fig_11}
\end{figure*}

\begin{figure*}[t!]
\centering
\includegraphics[width=1.0\textwidth]{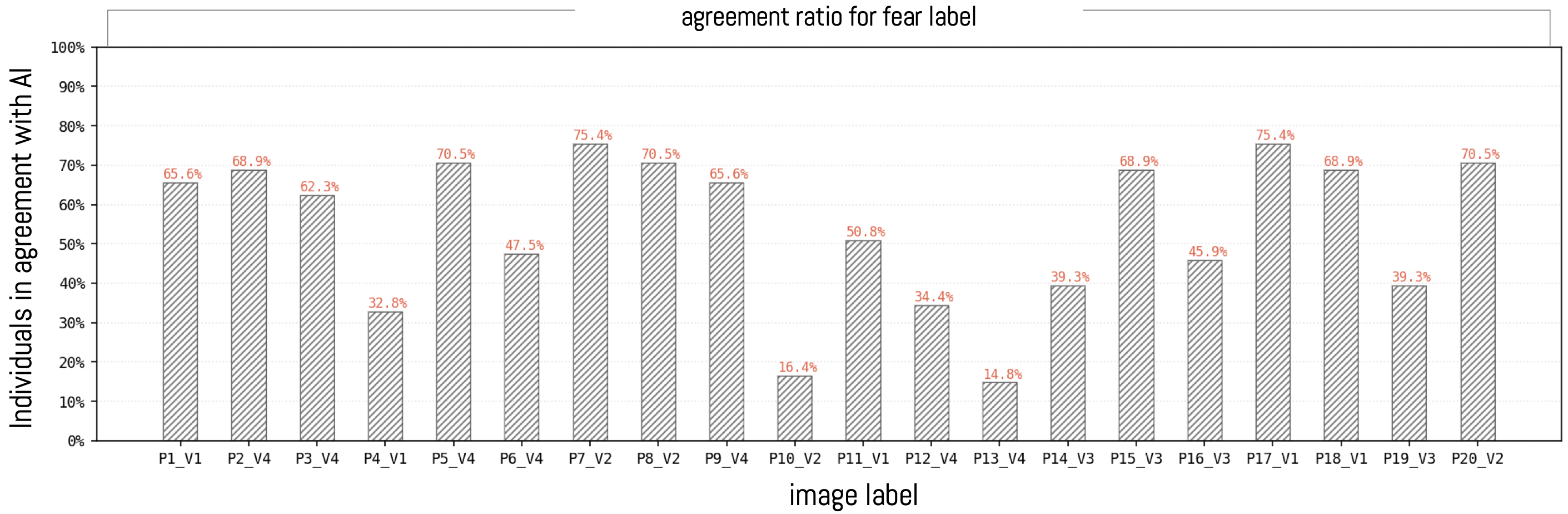}
\caption{Proportion of individuals who agree with the AI when classifying images labelled as Fear}\label{fig_12}
\end{figure*}

\begin{figure*}[t!]
\centering
\includegraphics[width=1.0\textwidth]{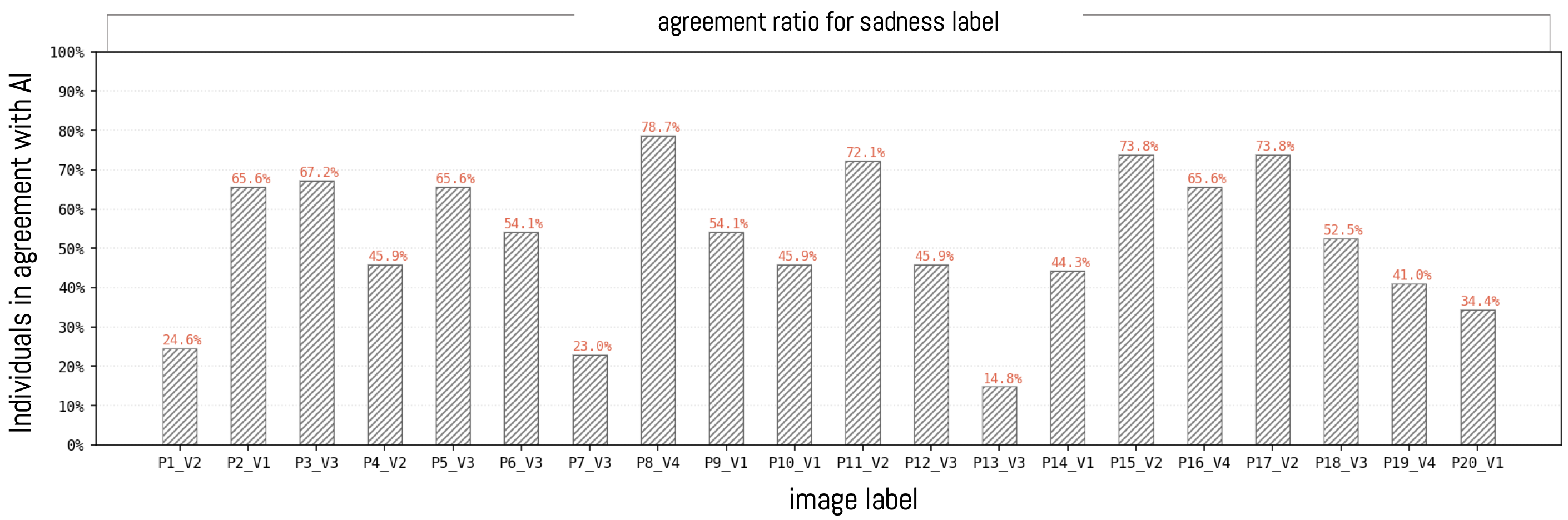}
\caption{Proportion of individuals who agree with the AI when classifying images labelled as Sadness}\label{fig_13}
\end{figure*}

In the case of the images that the AI has generated with the label ``Amusement'', as observed in Fig. \ref{fig_11}, it seems that the percentage of people who agree with the AI is, in general, lower than for the emotion ``Contentment'', reaching a proportion greater than 0.6 in only 3 of the 20 images. Although, at first glance, it may seem that the percentages do not differ so much among them, we find significant differences when comparing the proportions with Cochran's Test ($p$-value$<$0.001. See Table \ref{table_6}).

The emotion ``Fear'' seems to reach a higher agreement, since there are 11 images in which more than 60\% individuals have chosen the emotion with which they had been generated, reaching 75\% in two of them. However, we also find very low percentages (15\% and 16\%) for two of the images (Fig.{\ref{fig_12}). Again, there are significant differences when comparing the proportions ($p$-value$<$0.001. See Table \ref{table_6}).

\begin{table}[ht]
\caption{Comparison of proportion of agreement between AI and raters (mode) in the classification of images labelled with the same emotion (Cochran's test).}\label{table_6}%
\begin{tabular}{lllll}
\textbf{Emotion} & \textbf{Contentment} & \textbf{Amusement} & \textbf{Fear} & \textbf{Sadness} \\ \hline
\textit{$p$-value\footnotemark[1]} & \textless{}0.001*** & \textless{}0.001*** & \textless{}0.001*** & \textless{}0.001*** \\\hline
\end{tabular}
\footnotetext[1]{Significant differences are denoted with asterisks (*p\textless{}0.1, ** p\textless{}0.05, *** p\textless{}0.01).}
\end{table}

Finally, analyzing the emotion ``Sadness'', we observe in Fig. \ref{fig_13} that more than 60\% of participants agree with the AI when classifying 8 of the 20 images. As has happened with the other emotions, the differences between the proportions are statistically significant ($p$-value$<$0.001. See Table \ref{table_6}).

To sum, the results expressed in Table \ref{table_6} indicate that the proportion of people who agree with the generative tool is not similar for the different images, not even when we compare the classification of images generated for the same emotion ($p$-value$<$0.001). Analyzing the proportion of agreement when the emotion variable is dichotomous (negative/positive), there is also no similar percentage for images labelled with the same emotion (Table \ref{table_7}).

\begin{table}[ht]
\caption{Comparison of proportion of agreement between AI and raters (mode) in the classification of images labeled with the same negative or positive emotion (Cochran's test).}\label{table_7}
\begin{tabular}{lll}
\textbf{Emotion} & \textbf{Negative} & \textbf{Positive} \\ \hline
\textit{$p$-value\footnotemark[1]} & \textless{}0.001*** & \textless{}0.001*** \\ \hline
\end{tabular}
\footnotetext[1]{Significant differences are denoted with asterisks (*p\textless{}0.1, ** p\textless{}0.05, *** p\textless{}0.01).}
\end{table}

\section{Discussion}
The technological development of artificial intelligence has been exponential in recent years, and the advances in using this tool for analyzing emotional aspects across various fields of knowledge have been significant \cite{cao_comprehensive_2023}. However, to the best of our knowledge, we have not found studies that analyze the level of agreement between the emotions generated by generative artificial intelligence tools and human assessments. Closely related to this issue is the research conducted by Lopatovska (\cite{lopatovska_three_2016}), which focuses on works created by humans.

For this reason, we propose a novel methodology that begins with training a model using artworks catalogued by emotions from the Artemis dataset \cite{achlioptas_artemis_2021} to generate 20 landscape images. For each image, four emotional variants were created (contentment, amusement, sadness, and fear), which can be grouped into positive (contentment and amusement) and negative (sadness and fear) categories by dichotomizing the problem \cite{wang_research_2022,ionescu_computational_2022}, resulting in a total of 80 images. Using this dataset, an online questionnaire was designed to understand human emotional appreciation, obtaining 61 responses (33 male and 28 female) from participants across different countries, educational levels, and fields of study.

Using the obtained data, different analyses were conducted to address the research hypothesis regarding whether images created by generative processes with a specific emotion align with human emotional responses.

First, the agreement among evaluators regarding their emotional classifications of the AI-generated images was examined. For this purpose, Krippendorff's Alpha coefficient was utilized. According to Krippendorff \cite{krippendorff_reliability_2004}, the results indicate a fair level of agreement among the evaluators, with segmentation across the four emotional categories ($\alpha$=0.21–0.40). When emotions are categorized as positive and negative, the index increases, indicating a moderate level of agreement ($\alpha=$0.41–0.60).

A comparative analysis of different agreement indices indicates that Krippendorff's Alpha tends to yield low values \cite{eser_comparison_2022}. However, its utilization allows for a more nuanced interpretation of the phenomenon, accommodating more evaluators and various data typologies \cite{kim_interrater_2024}. The results obtained with Krippendorff's Alpha align with findings from other studies on agreement and emotions \cite{antoine_weighted_2014}, underscoring the high level of subjectivity in emotion classification. However, unlike our experiment, these studies utilize ordinal data. This level of subjectivity is further corroborated in our analysis using Cochran's Test, as it reveals that the evaluators do not achieve similar proportions when classifying the images, even within the same emotional category.

The following analysis aimed to evaluate the level of agreement for each emotion generated by the generative tool, considering this value as the true representation and comparing it with the mode of the classifications made by the observers.

An analysis of precision, recall, and F1 score reveals that the Fear category achieves the best classification performance across all analyzed groups. Specifically, within the Country segmentation, the Fear category exhibits an F1 score of $=0.92$. In contrast, certain emotional categories consistently demonstrate lower performance, particularly the Contentment category. This suggests a greater level of concordance between the expressions of the generative tool and user responses when the emotion is negative. When performing the same analysis with binarized emotions, the performance in both cases exceeds 90\% for the F1 score. This allows us to conclude that there is an agreement between human assignments and artificially generated images regarding the classification of an image as positive or negative.

Next, Fisher's Test was employed to determine precision and recall among groups of evaluators. Regarding the four emotion categories used, Sadness consistently yields the most significant differences in precision across all cases: in the male-female grouping ($p$-value$=$0.007), in the comparison between Technical Engineering and Social Sciences ($p$-value$=$0.07), and the education level comparison between undergraduate and postgraduate ($p$-value$=$0.042). In contrast, recall demonstrates statistically significant differences for the Fear emotion within the 'Engineering and Technology' knowledge area ($p$-value$=$0.064).

To further analyze the results, the Jaccard index was employed to measure how closely the evaluators' classifications align with the categories generated by the generative tool. The Chilean national group achieved the highest intersection ($J=0.7018$), followed by the group from the Social Sciences area ($J=0.7008$).

Finally, following recommendations to utilize more than one concordance index \cite{eser_comparison_2022}, Cohen's Kappa index was employed. Unlike Krippendorff's Alpha, Cohen's Kappa is limited to analysis involving two evaluators. This limitation was addressed in our research by using the mode and the categories generated by the tool.

Following Landis and Koch \cite{landis_measurement_1977}, the agreement results fall within the 'almost perfect' range across all groups when emotions are binarized into positive and negative. Among the four emotional categories, the Chilean nationality group achieved the highest agreement index, with a $k$=0.7833, followed by the group from the Engineering area ($k$=0.75). This analysis further indicates that agreement is more clearly achieved when the classification is simplified to positive and negative categories. In summary, it is observed that all indicators improve when the emotional classification problem is reduced to these two categories, a common approach in research on emotion recognition and study \cite{maithri_automated_2022}. This suggests that achieving agreement is more feasible both among evaluators and between the mode of human classification and the emotions generated by the generative tool. Notably, negative emotions yielded the highest levels of agreement in our study. Achieving complete agreement seems complex. Evidence for this is found in the research by Lopatovska \cite{lopatovska_three_2016}, which proposes three methodologies for the emotional classification of works of art created by humans, yet does not achieve a significant level of agreement in any case, even with human classifications.

\section{Limitations and future directions}
Among the main limitations of this research, the small number of evaluators who responded to the questionnaire stands out, as it constitutes a non-representative sample that hinders the ability to draw significant conclusions regarding the level of agreement on emotions, given their inherent subjectivity.

Additionally, it is recognized that social and cultural context plays a crucial role in emotional appreciation. Therefore, expanding the sample to include participants from more countries would facilitate comparative analyses. Similarly, involving individuals from a broader age range would enhance the comprehensiveness of the analysis.

Another factor to consider is that the generated sample for classification was limited to landscapes, which restricts the number of referential elements that could aid in classifying emotions more distinctly (e.g., faces). Future research should incorporate images with varying levels of representation and different elements, enabling an examination of the level of agreement across different degrees of representation. Furthermore, it would be interesting to investigate the key visual elements influencing emotional classification decisions, following previous research that has analyzed aspects such as color, shapes, and textures.

Finally, our study revealed that images conveying negative emotions were classified more effectively than those depicting positive emotions, suggesting that evaluators perceived negative emotions more clearly. This finding opens up new avenues for research to explore the underlying reasons for this phenomenon.

\section{Conclusion}

 Given the need to validate the content generated by artificial intelligence, this research focuses on emotional validation through statistical analysis of the level of agreement between a set of artificially generated images with associated emotions and the classification of these images by humans.

To achieve this, a methodology was proposed that includes training StyleGAN2-ADA using the Artemis dataset to generate 20 landscape images. For each image, four emotional variants were created (Contentment, Amusement, Fear, and Sadness), which can be grouped into positive and negative emotions. The human classification was conducted through an online questionnaire. Based on the obtained data, statistical analyses were performed to evaluate the level of agreement among individuals, the mode of the responses, the emotions generated by the AI, and to analyze proportions. 

The research conducted demonstrates the complexity involved in the study of emotions and the high level of subjectivity in their classification. Some results indicate, particularly with emotions binarized into positive and negative categories, a good level of agreement across different analyses, suggesting that the products generated by image tools appear to be reliable.

The main limitation of this research is the sample size, which cannot be considered representative. Future directions for this research include expanding the sample size, both in terms of the number of evaluators and their backgrounds, age, educational level, and areas of study.

We believe it is essential to advance in this field of study, as it would help validate the results generated by generative tools and enhance our understanding of their usefulness and limitations. Additionally, this research contributes to a deeper understanding of human emotional appreciation.

\bibliographystyle{unsrtnat}
\bibliography{references}  

\end{document}